\documentclass[lettersize,journal]{IEEEtran}
\usepackage{amsmath,amssymb,amsfonts}
\usepackage{algorithm}
\usepackage{array}
\usepackage{textcomp}
\usepackage{stfloats}
\usepackage{url}
\usepackage{verbatim}
\usepackage{graphicx}
\usepackage{cite}
\usepackage{amsbsy}
\usepackage{xcolor}
\usepackage[hidelinks]{hyperref}
\usepackage[noabbrev]{cleveref}
\usepackage{microtype}
\usepackage{relsize}
\usepackage{booktabs}
\usepackage{multirow}
\usepackage[position=bottom]{subfig}
\usepackage[font=small,labelfont=bf]{caption}
\usepackage{amsmath}
\usepackage{algorithm}
\usepackage[noend]{algpseudocode}

\usepackage{scalerel}
\usepackage{tikz}
\usetikzlibrary{svg.path}

\definecolor{orcidlogocol}{HTML}{A6CE39}
\tikzset{
  orcidlogo/.pic={
    \fill[orcidlogocol] svg{M256,128c0,70.7-57.3,128-128,128C57.3,256,0,198.7,0,128C0,57.3,57.3,0,128,0C198.7,0,256,57.3,256,128z};
    \fill[white] svg{M86.3,186.2H70.9V79.1h15.4v48.4V186.2z}
                 svg{M108.9,79.1h41.6c39.6,0,57,28.3,57,53.6c0,27.5-21.5,53.6-56.8,53.6h-41.8V79.1z M124.3,172.4h24.5c34.9,0,42.9-26.5,42.9-39.7c0-21.5-13.7-39.7-43.7-39.7h-23.7V172.4z}
                 svg{M88.7,56.8c0,5.5-4.5,10.1-10.1,10.1c-5.6,0-10.1-4.6-10.1-10.1c0-5.6,4.5-10.1,10.1-10.1C84.2,46.7,88.7,51.3,88.7,56.8z};
  }
}

\newcommand\orcidid[1]{\href{https://orcid.org/#1}{\mbox{\scalerel*{
    \begin{tikzpicture}[yscale=-1,transform shape]
        \pic{orcidlogo};
    \end{tikzpicture}
}{|}}}}

\makeatletter
\def\BState{\State\hskip-\ALG@thistlm}
\makeatother
\begin{document}

\title{MixNet: Joining Force of Classical and Modern Approaches toward The Comprehensive Pipeline in Motor Imagery EEG Classification}
\author{
     Phairot~Autthasan$^{\orcidid{0000-0002-9566-8382}}$, Rattanaphon~Chaisaen$^{\orcidid{0000-0003-1521-9956}}$, 
     Huy~Phan$^{\orcidid{0000-0003-4096-785X}}$,  Maarten~De Vos$^{\orcidid{0000-0002-3482-5145}}$, and Theerawit~Wilaiprasitporn$^{\orcidid{0000-0003-4941-4354}}$ \vspace{-0.1in}
    \thanks{This work was supported by PTT Public Company Limited, The SCB Public Company Limited, Thailand and National Research Council of Thailand (N41A640131) \textit{(Phairot Autthasan and Rattanaphon Chaisaen contributed equally to this work)} \textit{($^{*}$Corresponding author: Theerawit Wilaiprasitporn).}}
    \thanks{P. Autthasan, R. Chaisaen, and T. Wilaiprasitporn are with Bio-inspired Robotics and Neural Engineering (BRAIN) Lab, School of Information Science and Technology (IST), Vidyasirimedhi\break Institute of Science \& Technology (VISTEC), Rayong, 21210, Thailand (e-mail: theerawit.w@vistec.ac.th).}
    \thanks{H. Phan is with Amazon AGI, Cambridge, MA 02142 USA. The work does not relate to H.P.’s position at Amazon.}
    \thanks{M. De Vos is with the Department of Engineering and with the Department of Development and Regeneration, KU Leuven, Leuven, Belgium, 3001.}
    \thanks{Code examples, and other supporting materials are available on \break\url{https://github.com/Max-Phairot-A/MixNet}}} 
    
\maketitle

\begin{abstract}

Recent advances in deep learning (DL) have significantly impacted motor imagery (MI)-based brain-computer interface (BCI) systems, enhancing the decoding of electroencephalography (EEG) signals. However, most studies struggle to identify discriminative patterns across subjects during MI tasks, limiting MI classification performance. In this paper, we propose MixNet, a novel classification framework designed to overcome this limitation by utilizing spectral-spatial signals from MI data, along with a multi-task learning architecture named MIN2Net, for classification. Here, the spectral-spatial signals are generated using the filter-bank common spatial patterns (FBCSP) method on MI data. Since the multi-task learning architecture is used for the classification task, the learning in each task may exhibit different generalization rates and potential overfitting across tasks. To address this issue, we implement adaptive gradient blending, simultaneously regulating multiple loss weights and adjusting the learning pace for each task based on its generalization/overfitting tendencies. Experimental results on six benchmark datasets of different data sizes demonstrate that MixNet consistently outperforms all state-of-the-art algorithms in subject-dependent and -independent settings. Finally, the low-density EEG-MI classification results show MixNet's superiority over state-of-the-art algorithms, offering promising implications for Internet of Thing (IoT) applications such as lightweight and portable EEG wearable devices based on low-density montages.
\end{abstract}


\section{Introduction}
\label{sec:introduction}

\IEEEPARstart{B}{RAIN}-computer interfaces (BCIs) are a transformative technology that establishes a direct communication channel between the human brain and external devices \cite{WOLPAW2002767}. Among the numerous neural acquisition techniques used in BCIs, electroencephalography (EEG) has attracted considerable interest due to its non-invasiveness, affordability, and high temporal resolution\cite{1454811, 95357}. Motor imagery (MI)-based BCI systems are increasingly popular in BCI paradigms due to their advantage of not requiring external stimuli\cite{5165082, 7802578}. Within such an MI-based BCI system, participants are instructed to mentally imagine the movement of different body parts, evoking neural activity in specific cerebral regions associated with those movements. Throughout the MI process, the co-occurrence of event-related desynchronization (ERD) and event-related synchronization (ERS) patterns in EEG signals reflects event-related modulations in brain activity within particular frequency bands, such as mu (9–13 Hz) and beta (22–29 Hz), over the motor cortex region\cite{NAM2011567, ZICH2015438}. The ERD/ERS patterns generated by imagined movements are matched with actual movements and then decoded to determine the user’s intent. Furthermore, the output signal can be used to control external devices\cite{ima20283}.

Nevertheless, MI-based BCI systems encounter notable challenges. Firstly, the EEG signals have a low signal-to-noise ratio (SNR), making them highly sensitive to noise interference\cite{847807, 7134704}. Another limitation is the considerable variability observed in EEG signals across different subjects, posing a substantial challenge to accurately decoding these oscillatory neural activities. Recently, researchers have developed calibration-free or subject-independent methods to tackle the issue of EEG data variability across subjects\cite{8897723, min2net}. These methods are trained and tested using data from different groups of subjects, enabling new users to utilize the BCI system without needing a calibration phase. For the MI-based BCI system to help capture and learn generalized MI-EEG features across subjects, it is essential to develop powerful feature extraction and classification algorithms. Deep learning (DL) has recently emerged as an up-and-coming technique and demonstrated remarkable success in numerous fields, including computer vision, bio-signals, speech recognition, and natural language processing\cite{9325918}. In contrast to conventional methods, which rely on manually hand-crafted features, DL models can learn complex patterns directly from multi-dimensional data. This capability has attracted numerous researchers' interest in BCIs, resulting in the development of advanced DL architectures that have demonstrated significant improvements in EEG-based motor imagery (MI) classification\cite{ALSAEGH2021102172}.

Convolutional neural networks (CNNs) have been extensively applied in EEG-MI classification due to their ability to learn temporal and spatial features from EEG signals effectively \cite{tonio_paper, 8310961, Lawhern_2018}. By leveraging the power of 2D-CNNs, these models can capture local and global connectivity patterns within EEG data. The combination of CNNs and long short-term memory units (LSTMs) has been extensively employed to capture long-term dependencies in temporal features\cite{8745473, 8556024}. This fusion permits extracting relevant temporal and spatial features, enhancing the overall performance of EEG-MI classification. Despite the notable successes of existing deep learning approaches in decoding EEG signals in various MI datasets, their performance is limited to the subject-dependent task, where the methods are trained and tested using data from the same subject. In other words, these models have difficulty adapting and performing well when evaluated using data from unseen users. This limitation hinders their applicability and potential real-world deployment in BCI systems.

In recent years, our previous work \cite{min2net} introduced a deep learning architecture known as MIN2Net, explicitly developed to capture generalized MI-EEG features across subjects in the subject-independent task. This architecture demonstrated acceptable performance in subject-independent learning over a large-scale dataset. However, its effectiveness was compromised when attempting to recognize generalized EEG-MI features with limited training data, such as using a small data size and training data from only a single subject in subject-dependent learning. Another drawback is incorporating multi-task learning into MIN2Net (e.g., unsupervised, supervised, and deep metric learning) for EEG-MI classification proved challenging, given the differing generalization rates and potential overfitting across tasks. MIN2Net was the reliance on an exhaustive parameter search to identify the optimal set of loss weights. This time-consuming process yielded varying optimal weight configurations when applied to different datasets.

This paper proposes MixNet, a novel framework that builds upon the foundational principles of MIN2Net, addressing limitations in network learning with small data sizes and time-consuming loss weights optimization. MixNet utilizes filter-bank common spatial patterns (FBCSP) for data pre-processing, extracting discriminative patterns from EEG-MI data. This results in spatially filtered or spectral-spatial signals, preserving meaningful information across EEG classes. We adjusted and adopted the adaptive gradient blending concept, previously applied in sleep staging tasks by Huy et al. \cite{adaptive_gradient_huy}, which is incorporated to regulate multiple loss weights concurrently. This approach alleviates the need for manual parameter tuning, leading to a more efficient and effective optimization process. The proposed method achieves remarkable performance in subject-dependent and subject-independent MI classification tasks by leveraging the spectral-spatial signals and employing an adaptive gradient blending approach for loss weight optimization.

Additionally, MixNet outperforms state-of-the-art algorithms in handling MI classification on low-density EEG-MI signals. As the MixNet developed for improving low- and high-density EEG classification, this advancement has promising implications for Internet of Thing (IoT) applications such as the EEG classification on EEG signals that are from light-weight and portable EEG wearable devices based on low-density montages; most of these devices are more comfortable for users, reducing the effort to set up and also expanding their usage in the real world, including stroke identification \cite{GOTTLIBE202021} and robot control \cite{10230307}.

The remainder of this paper is organized as follows. Section II presents an overview of related work on MI classification topics. Section III details the pre-processing steps applied to EEG data and outlines the architecture of the proposed method. Section IV presents the experimental results, indicating the performance of the proposed method. Section V provides a comprehensive discussion of the experimental results and their implications. Finally, the main findings and potential avenues for future research are summarized in Section VI.

\section{Related work}
This section thoroughly reviews the evolution and progress in EEG-based MI classification. Subsequently, we address the existing limitations in the current research on MI-based BCI. Furthermore, we provide a detailed exposition of the concepts of autoencoders (AE) and deep metric learning (DML), both of which bear relevance to our study. Lastly, we outline the rationale behind undertaking this research endeavor.
\begin{figure*}[ht]
  \centering
  \subfloat[\label{fig:input_data}]{\includegraphics[width=1.3\columnwidth]{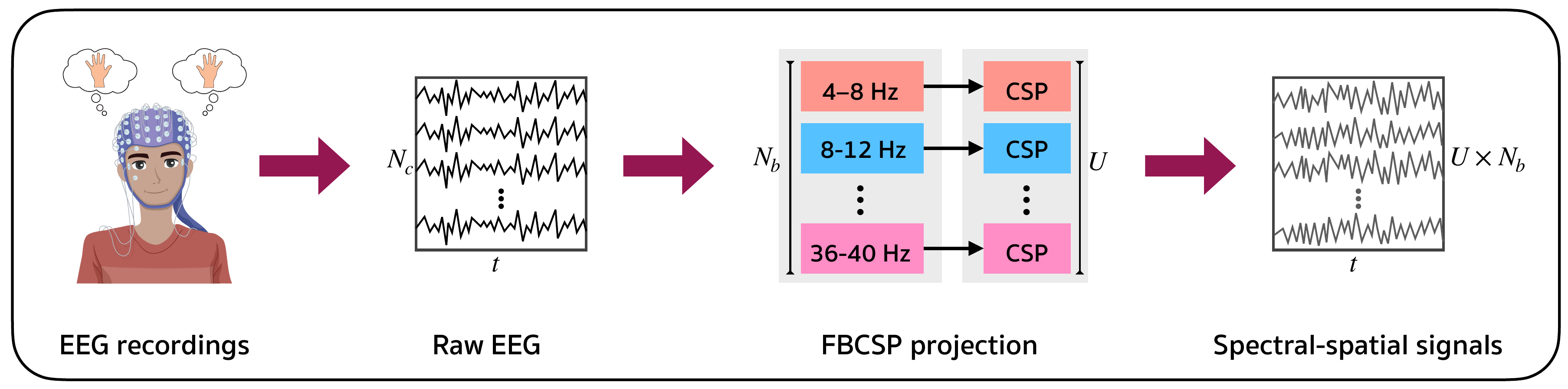}} \\
  \subfloat[\label{fig:model}]{\includegraphics[width=1.3\columnwidth]{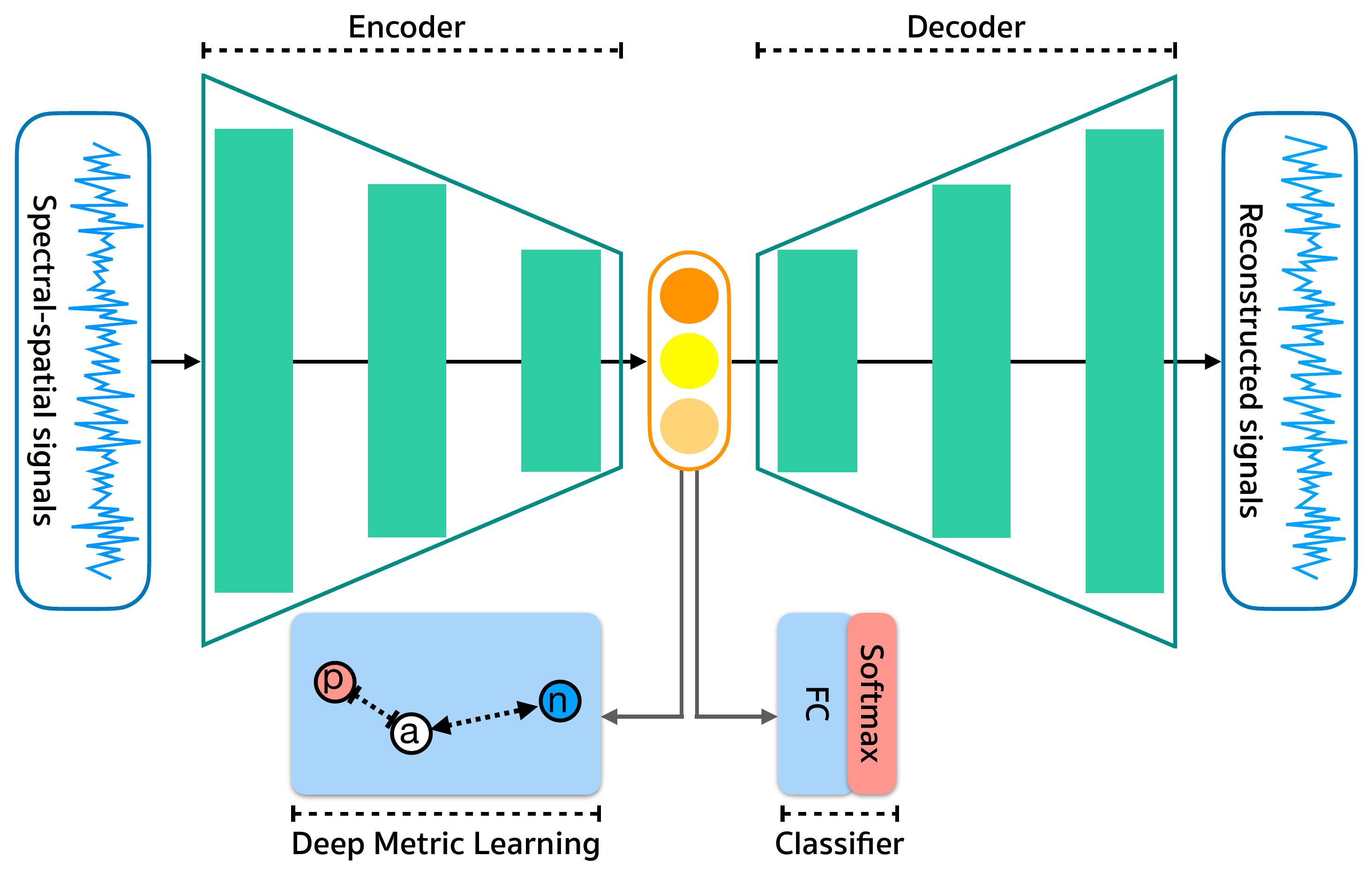}}
  \caption{Overall visualization of MixNet framework. (a) exhibits an overview of the preparation process for spectral-spatial signals. (b) demonstrates the multi-task autoencoder (AE) of MixNet, designed to simultaneously address three tasks: autoencoder, deep metric learning, and supervised learning. This framework is employed for EEG-MI classification. More in-depth insights into the network architecture can be found in \autoref{tab:model}.}
\end{figure*}
\subsection{The Progressive Development of MI based on BCI}
The progressive evolution of machine learning has inspired BCI researchers to propose more advanced intelligent algorithms employing subject-dependent paradigms to improve motor imagery (MI) decoding performance based on EEG signals. Common spatial pattern (CSP) is one of the most prominent approaches widely used for feature extraction in MI-based BCI\cite{895946}. CSP efficiently extracts discriminative features by maximizing the variance differences between the different classes of EEG signals. Moreover, several methods are built based on extensions of CSP, such as common spatiospectral pattern (CSSP)\cite{1495698}, filter bank common spatial pattern (FBCSP)\cite{4634130}, and bayesian spatio-spectral filter optimization (BSSFO)\cite{6175024}.

Among advanced CSP algorithms, FBCSP is the most widely used, which extends the approach by incorporating multiple frequency bands instead of being limited to a specific band. It has demonstrated state-of-the-art performance in EEG-based MI classification as seen in \cite{second_fbcsp}. Upon processing EEG signals through FBCSP, meaningful brain features are extracted. Subsequently, a feature selection method such as mutual information-based best individual feature (MIBIF)\cite{4634130} is employed to identify the most discriminative features. In the realm of classification, numerous conventional algorithms, such as support vector machine (SVM) and linear discriminant analysis (LDA), have been extensively utilized to decode the extracted features, as seen in \cite{8897723, second_fbcsp}. Furthermore, several algorithms have extended the CSP paradigm, promising performance improvements\cite{5593210, 8353425, 9349966}. Despite the significant achievements of existing CSP and its modified algorithms in EEG decoding for various Motor Imagery (MI) datasets, it is evident that their efficacy is primarily restricted to the subject-specific task. Regrettably, their effectiveness in the subject-independent task is a topic of ongoing research and requires additional refinement to improve their performance and generalizability.

Recently, advanced deep learning techniques \cite{8897723, ZHANG20211, min2net, nagarajan2023relevance} have demonstrated the ability to effectively utilize large-scale EEG data and knowledge from analogous or related subjects/sessions, aiming to reduce the need for calibration. In a more recent paper, Kwon \textit{et al.}\cite{8897723} proposed a subject-independent framework based on CNN architectures that incorporate spectral-spatial feature representation to enhance subject-independent MI classification, achieving state-of-the-art performance on the OpenBMI dataset. Subsequently, in our previous work \cite{min2net}, we introduced MIN2Net, a novel end-to-end neural network architecture based on a multi-task autoencoder, which achieved state-of-the-art performance on the SMR-BCI and OpenBMI datasets in the subject-independent setting. However, MIN2Net's performance was limited in identifying generalized EEG-MI features under conditions of limited training data, such as with small dataset sizes or when relying on training data from a single subject in subject-dependent learning.

\subsection{Deep Metric Learning in MI based on BCI}
Deep metric learning (DML) is an advanced and effective technique based on the fundamental concept of distance metrics\cite{8103121}. Its primary objective is to acquire highly informative data representations that enable precise data similarity measurement. This is accomplished using embedded features obtained from a metric learning network. Classical similarity metric functions, such as Euclidean distance, Mahalanobis distance, and Cosine distance, are explicitly engaged as the distance metric between pairs of data points within the DML framework.

In recent years, there has been significant advancement in developing several loss functions designed explicitly for DML. These loss functions are created with the primary objective of improving the discrimination of features. Several loss functions have been identified as significant in this context, including contrastive loss\cite{1640964}, triplet loss\cite{7298682}, quadruplet loss\cite{quadruplet_paper}, and multi-similarity loss\cite{wang2019multi}. By adeptly utilizing these loss functions, similarity measures are computed on correlated samples, compelling samples belonging to the same class to converge while concurrently inducing samples from distinct classes to diverge. One notable differentiation between DML losses and traditional loss functions, such as cross-entropy loss, is their dependency on contrastive pairs, triplets, or quadruplets of data for gradient determination. 

More recently, DML has been extensively utilized in EEG-BCI research, providing novel opportunities for decoding human brain activity more precisely and efficiently \cite{8257016, 9116935, min2net, 9837422}. Specifically, integrating the DML technique into advanced deep learning architectures, such as autoencoders and convolutional neural networks (CNNs), enables the model to automatically learn and extract meaningful features from high-dimensional embeddings and optimize the entire system for the task of capturing similarities and differences between EEG classes. The combination of deep metric learning in multi-task autoencoder-based CNN architectures indicates great promise for advancing EEG-BCI technology and has demonstrated promising results in addressing the long-standing problem of inter-subject variability, as seen in our previous work \cite{min2net}.


\subsection{Autoencoders in MI based on BCI}
The Autoencoder (AE) concept, a widely recognized component in unsupervised learning algorithms, was first introduced in 1986\cite{fisrt_ae}. Autoencoders (AEs) have been widely used and have demonstrated their versatility in various domains, such as data compression, denoising, dimensionality reduction, and feature extraction\cite{Hinton504, 552239, SAE_paper}. The AE network architecture can efficiently extract crucial features from labeled and unlabeled input data, forming a latent representation. The latent representation plays a fundamental role in the later reconstruction of the original input data. The training objective for this network architecture is centered on reducing the reconstruction loss corresponding to the input data. Moreover, significant research findings suggest that the efficacy of the obtained latent representation is much enhanced when the reconstructed data closely mirrors the prominent features of the original input data.

One such advanced AE architecture is the denoising sparse autoencoder (DSAE) \cite{8429921}, which is proposed to improve EEG-based epileptic seizure detection. The main objective of DSAE is to improve the accuracy of epileptic seizure detection by efficiently reconstructing the original EEG signals from corrupted inputs. The sparsity constraint integrated into the DSAE is crucial in achieving this efficiency, offering significant potential for more reliable and precise results. Another area where AE-based methods have shown remarkable potential is in handling biopotentials and telemonitoring systems. A technique based on compressed sensing (CS) and AE \cite{7752836} has demonstrated outstanding results in data compression and accurate classification of electrocardiogram (ECG) and EEG signals. 

In more recent papers, the utilization of DL-based AE architectures has grown in popularity. In the works of Parashiva \textit{et al.}\cite{parashiva2019new} and Mammone \textit{et al.}\cite{mammone2023autoencoder}, researchers have shown this trend through integrating AEs with FBCSP technique. These studies have proven that combining AEs and FBCSP can extract crucial features, resulting in excellent EEG-based MI classification results. One limitation of these approaches is that they do not offer end-to-end learning paradigms. Even though latent representations were successfully compressed during the encoding-decoding process, further classification is still required to accurately distinguish and classify various classes or types of EEG patterns.

In our previous work \cite{min2net}, we introduced MIN2Net, a novel end-to-end neural network architecture based on a multi-task autoencoder. MIN2Net simultaneously learns deep features from unsupervised EEG-MI reconstruction and supervised EEG-MI classification. It demonstrated excellent performance in extracting discriminative features for MI decoding. However, the model cannot extract discriminative patterns from raw EEG signals when the number of training samples is relatively scarce. Therefore, from all angles, this study aims to develop an enhanced version of MIN2Net capable of achieving high performance regardless of the size of the training dataset.

\section{Methods}

\begin{table}[ht]
\caption{Description of all benchmark datasets}
\label{tab:data_description}
\centering
\resizebox{1\columnwidth}{!}{%
    \begin{tabular}{@{}cccccc@{}}
    \toprule[0.2em]
    \textbf{Datasets} & \textbf{\# Subjects} & \textbf{MI Task}          & \textbf{\# Channels} & \textbf{Selected Channel location}                                                                                                                                                                              & \textbf{Frequency (Hz.)} \\ \midrule[0.1em]
    BCIC IV 2a        & 9                    & \begin{tabular}[c]{@{}c@{}}Left hand \\ vs. Right hand \end{tabular}                        & 20                   & \begin{tabular}[c]{@{}c@{}}$FC_3$, $FC_1$, $FCz$, $FC_2$, $FC_4$, \\ $C_5$, $C_3$, $C_1$, $Cz$, $C_2$, \\ $C_4$, $C_6$, $CP_3$, $CP_1$, $CPz$, \\ $CP_2$, $CP_4$, $P_1$, $Pz$, $P_2$\end{tabular}      & 250                               \\ \midrule[0.1em]
    BCIC IV 2b        & 9                    & \begin{tabular}[c]{@{}c@{}}Left hand \\ vs. Right hand \end{tabular}                         & 3                   & \begin{tabular}[c]{@{}c@{}}$C_3$, $Cz$, $C_4$\end{tabular}      & 250                               \\ \midrule[0.1em]
    BNCI2015-001        & 12                    & \begin{tabular}[c]{@{}c@{}}Right hand \\ vs. Feet \end{tabular}                         & 13                   & \begin{tabular}[c]{@{}c@{}}$FC_3$, $FCz$, $FC_4$, $C_5$, $C_3$, \\ $C_1$, $Cz$, $C_2$, $C_4$, $C_6$, \\ $CP_3$, $CPz$, $CP_4$\end{tabular}      & 512                               \\ \midrule[0.1em]
    SMR-BCI           & 14                   & \begin{tabular}[c]{@{}c@{}}Right hand \\ vs. Feet \end{tabular}                              & 15                   & \begin{tabular}[c]{@{}c@{}}Small Laplacian electrode were\\placed at $C_3$, $Cz$, and $C_4$.\\Distances between neighboring\\electrodes were 2.5 cm. \end{tabular}                                                                                                                                                                                                      & 512                               \\ \midrule[0.1em]
    High-Gamma           & 14                   & \begin{tabular}[c]{@{}c@{}}Left hand \\ vs. Right hand \end{tabular}                          & 20                   & \begin{tabular}[c]{@{}c@{}}$FC_5$, $FC_3$, $FC_1$, $FC_2$, $FC_4$, \\ $FC_6$, $C_5$, $C_3$, $C_1$, $Cz$, \\ $C_2$, $C_4$, $C_6$, $CP_5$, $CP_3$, \\ $CP_1$, $CPz$, $CP_2$, $CP_4$, $CP_6$\end{tabular} & 500                            \\ \midrule[0.1em]
    OpenBMI           & 54                   & \begin{tabular}[c]{@{}c@{}}Left hand \\ vs. Right hand \end{tabular}                         & 20                   & \begin{tabular}[c]{@{}c@{}}$FC_5$, $FC_3$, $FC_1$, $FC_2$, $FC_4$, \\ $FC_6$, $C_5$, $C_3$, $C_1$, $Cz$, \\ $C_2$, $C_4$, $C_6$, $CP_5$, $CP_3$, \\ $CP_1$, $CPz$, $CP_2$, $CP_4$, $CP_6$\end{tabular} & 1,000                             \\ \bottomrule[0.2em]
    \end{tabular}
}
\end{table}

\begin{table*}[ht]
\caption{MixNet's architecture, where $U$ is the number of spatial filters, $t$ is the number of time points, $N_b$ is the number of frequency bands, $z$ is the size of latent vector and $N_{class}$ is the number of classes. Noted that the data format of Conv2D is ``channels last''}
\label{tab:model}
\centering
\resizebox{1.4\columnwidth}{!}{%
\begin{tabular}{@{}lllllllll@{}}
\toprule[0.2em]
\multicolumn{2}{l}{\textbf{Blocks}}                      & \multicolumn{1}{l}{\textbf{Layer}} & \textbf{Filter} & \multicolumn{1}{l}{\textbf{Size}} & \multicolumn{1}{l}{\textbf{Stride}} & \multicolumn{1}{l}{\textbf{Activation}} & \multicolumn{1}{l}{\textbf{Options}} & \multicolumn{1}{l}{\textbf{Output}} \\ \midrule[0.1em] 
\multirow{13}{*}{Autoencoder} & \multirow{8}{*}{Encoder} & Input                              &                 &              (1, $t$, $U\times N_{b}$)                      &                                     &                                         &                                      & (1, $t$, $U\times N_{b}$)                            \\
                              &                          & Conv2D                             & $U\times N_{b}$               & (1, 64)                            & 1             & ELU                                     & padding=same                         & (1, $t$, $U\times N_{b}$)                            \\
                              &                          & BatchNormalization                          &                 &                                   &                                     &                                         &                                      & (1, $t$, $U\times N_{b}$)                            \\
                              &                          & AveragePooling2D                      &                 & (1, $t$/100)                       &                                     &                                         &                                      & (1, 100, $U\times N_{b}$)                           \\
                              &                          & Conv2D                             & $(U\times N_{b})/2$              & (1, 32)                            & 1              & ELU                                     & padding=same                         & (1, 100, $U\times N_{b}/2$)                          \\
                              &                          & BatchNormalization                                 &                 &                                   &                                     &                                         &                                      & (1, 100, $U\times N_{b}/2$)                          \\
                              &                          & AveragePooling2D                   &                 & (1, 4)                             &                                     &                                         &                                      & (1, 25, $U\times N_{b}/2$)                           \\
                              &                          & Flatten                            &                 &                                   &                                     &                                         &                                      & (25 $\times$ $U\times N_{b}/2$)                               \\ \cmidrule[0.1em](l){2-9} 
                              & Latent                   & FC                                 &                 & ($z$)                               &                                     &                                         &                                      & (\textit{z})                                 \\ \cmidrule[0.1em](l){2-9} 
                              & \multirow{4}{*}{Decoder} & FC                                 &                 & (25 $\times$ $U\times N_{b}/2$)                             &                                     &                                         &                                      & (25 $\times$ $U\times N_{b}/2$)                               \\
                              &                          & Reshape                            &                 & (1, 25, $U\times N_{b}/2$)                       &                                     &                                         &                                      & (1, 25, $U\times N_{b}/2$)                         \\
                              &                          & Conv2DTranspose                    & $U\times N_{b}/2$              & (1, 64)                            & 4               & ELU                                     & padding=same                         & (1, 100, $U\times N_{b}/2$)                        \\
                              &                          & Conv2DTranspose                    & $U\times N_{b}$               & (1, 32)                            & $t$/100                              & ELU                                     & padding=same                         & (1, $t$, $U\times N_{b}$)                            \\ \midrule[0.1em] 
\multicolumn{2}{c}{\multirow{1}{*}{Deep Metric learning}}     & Latent                          &                 &                                  &                                     &                                         &                                      & ($z$)                                 \\ \midrule[0.1em]
\multicolumn{2}{c}{\multirow{2}{*}{Supervised Learning}}          & Latent                          &                 &                                   &                                     &                                         &                                      & (\textit{z})                                 \\
\multicolumn{2}{c}{}                                     & FC                                 & $N_{class}$             &                                   &                                     & softmax                                 &                                      & ($N_{class}$)                                 \\ \bottomrule[0.2em]
\end{tabular}
}
\end{table*}

\subsection{Databases}
\label{databases}

The proposed method, along with all baseline methods, was assessed using various datasets, including BCIC IV 2a\cite{BCIC_IV_2a}, BCIC IV 2b\cite{BCIC_IV_2b}, BNCI2015-001\cite{faller2012autocalibration}, SMR-BCI\cite{BCIC_SMR}, High-Gamma\cite{tonio_paper}, and OpenBMI\cite{lee_dataset} datasets. The BCIC IV 2a, BCIC IV 2b, BNCI2015-001, and SMR-BCI datasets are widely recognized as benchmark datasets for MI classification, originating from Graz University of Technology. The High-Gamma dataset, which has been made available by the University of Freiburg, is recognized as the second largest MI dataset. Lastly, the evaluation also included the OpenBMI dataset, which is the largest public MI dataset to date, provided by Korea University.

In the high-density EEG motor imagery (EEG-MI) classification task, we considered all EEG channels from the BCIC IV 2a, BNCI2015-001, and SMR-BCI datasets, as these datasets directly provide comprehensive EEG channels associated with the motor cortex area. Following previous studies \cite{8897723, min2net}, we selected 20 EEG channels in the motor cortex region from the High-Gamma and OpenBMI datasets for our analysis. We utilized the BCIC IV 2b dataset for the low-density EEG-MI classification task, which provides EEG data from 3 channels ($C_3$, $Cz$, and $C_4$) placed over the motor cortex region. Further details about these databases are provided in \autoref{tab:data_description}.

Additionally, all EEG data in the considered datasets were downsampled to a sampling frequency of 100 Hz, and the MI period for all datasets was selected as the time interval between 0 s and 4 s after stimulus onset. The EEG signals utilized in this study are time-domain signals with a duration of four seconds.

\subsection{Generation of Spectral-spatial Signals}

In this study, the raw EEG signals are time-domain signals that change over time with 4 seconds long. Formally, we denote $X_{i}\in\mathbb{R}^{N_{c}\times t}$ as a single-trial EEG data from the $i^{th}$ trial and let $y_{i}\in\{1,2,...,N_{class}\}$ denote its corresponding label, where $N_{c}$ is the number of channels, $t$ is the number of sampled time points, and $N_{class}$ is the number of classes. 

Using a fifth-order Butterworth bandpass filter, a filter bank was used to decompose the single-trial EEG data $X$ into multiple frequency bands. Here, the filter bank is defined as $B = \{b_{k}\}_{k=1}^{N_{b}} \in\{[4,8],[8,12],...,[36,40]\}$, where $N_{b}$ is the number of frequency bands, is predefined from 4 to 40 Hz. Each bandpass filter has a range of 4 Hz. While some variations in the filter bank configurations are effective, these specific bandpass frequency ranges are used due to their ability to provide stable frequency response, spanning the range of 4 -- 40 Hz, as observed in the work by\cite{second_fbcsp}. Filtered EEG signals will be denoted as $E_{{b_{k}},i}\in\mathbb{R}^{N_{c}\times t}$.

After filtering the EEG signals by the filter bank, the CSP algorithm was used for spatial filtering\cite{4634130, second_fbcsp}. The CSP algorithm is well-known for its efficacy in estimating and applying a projection matrix to transform EEG signals linearly, maximizing the variance of signals belonging to one class and minimizing the variance of signals belonging to the other. The spatial filtering procedure involves linearly transforming EEG data using the CSP algorithm as follows:
\begin{equation} \label{first_csp}
 Z_{{b_{k}},i} = W_{b_{k}}^T E_{{b_{k}},i},
\end{equation}
Where $Z_{{b_{k}},i}\in\mathbb{R}^{N_{c}\times t}$ denotes $E_{{b_{k}},i}$ after spatial filtering, $W_{b_{k}}\in\mathbb{R}^{N_{c}\times {N_{c}}}$ denotes the CSP projection
matrix; $T$ denotes the transpose operator.

The CSP algorithm computes the transformation matrix $W_{b_{k}}$ through the solution of the eigenvalue decomposition problem as shown in \autoref{second_csp}, thereby producing features with optimized variances to discriminate between the two classes of EEG signals effectively:
\begin{equation} \label{second_csp}
 \Sigma_{{b_{k}},1} W_{b_{k}} = (\Sigma_{{b_{k}},1} + \Sigma_{{b_{k}},2}) W_{b_{k}} D_{b_{k}},
\end{equation}
Where $\Sigma_{{b_{k}},1}$ and $\Sigma_{{b_{k}},2}$ are the estimation of the covariance matrices of the filtered EEG signals for motor imagery class 1 and class 2, respectively. $D_{b_{k}}$ is the diagonal matrix that contains the eigenvalues of $\Sigma_{{b_{k}},1}$. By considering all the filtered EEG trials $N_{i}$, the covariance matrix for each class of the $b_{k}$ filtered EEGs is then given by:

\begin{equation} \label{covariance matrix}
\Sigma_{{b_{k}},y} = \frac{1}{N_{i}} \sum_{i=1}^{N_{i}} \frac{E_{{b_{k}},i}^{(y)} E_{{b_{k}},i}^{(y)T}}{tr(E_{{b_{k}},i}^{(y)} E_{{b_{k}},i}^{(y)T})}
\end{equation}

The spatially filtered EEG signal or spectral-spatial EEG signal $Z_{{b_{k}},i}$ in \autoref{first_csp} using first and last $\frac{U}{2}$ columns of spatial filter $W_{b_{k}}$ from \autoref{second_csp} can maximize the differences in the variance of the two classes of filtered EEG signals, thus denoted as $\bar{W}_{b_{k}}$. For example, the number of spatial filters $U = 4$ means that the two largest and the two smallest eigenvalues of the spatial filter $W_{b_{k}}$ are selected. The spectral-spatial signals are finally computed as follows: 
\begin{equation} \label{last_csp}
 \bar{Z}_{{b_{k}},i} = \bar{W}_{b_{k}}^T E_{{b_{k}},i},
\end{equation}
Where $\bar{Z}_{{b_{k}},i}\in\mathbb{R}^{U\times t}$ and $\bar{W}_{b_{k}}\in\mathbb{R}^{N_{c}\times U}$.

\subsection{Classification With MixNet}
In this study, the proposed method builds upon the
the idea of our previous paper presented MIN2Net and still keeps the network architecture the same as the original MIN2Net. MixNet consists of three main modules: autoencoder, deep metric learning, and supervised learning, and its architecture is optimized by minimizing three different loss functions simultaneously: reconstruction, cross-entropy, and triplet loss functions. 

\subsubsection{Autoencoder}
The autoencoder module, which is a part of the MixNet framework, has two main components: the encoder, expressed as $z = q(x)$, and the decoder, represented as $\hat{x} = p(z)$. The input signal $x$ is subjected to encoding within the encoder component, transforming $x$ into a latent vector $z$ via dimensionality reduction. In contrast, the decoder module receives a given latent vector $z$ and continues to decode it, resulting in the reconstruction of the input signal $\hat{x}$. The primary objective of the AE module is to learn how to map data into a latent space that retains pertinent information for reconstructing the data. By utilizing spectral-spatial EEG signals from FBCSP as inputs for the AE module in this study, the AE can effectively recognize instances and preserve instances' discriminative patterns corresponding to different classes.

The encoder module consists of two separate CNN blocks. Each block is built of a Conv2D layer, followed by a batch normalization (BN) layer, an exponential linear unit (ELU), and an average pooling layer (AveragePooling2D). The final output of the last CNN block serves as the input to a fully connected (FC) layer. This FC layer aids in the process of mapping the latent representation. The decoder module has a symmetrical construction to that of the encoder module. The latent vector is passed via an FC layer to preserve consistency in the input dimensions of the CNN blocks. It then proceeds to a reshape layer that arranges the data into a suitable dimension. The decoder module consists of two CNN blocks, each using a Conv2DTranspose layer with a stride value of 4. Subsequently, an ELU layer is used to achieve its intended purpose. More details on the dimensions of the input size at each layer in the architecture of MixNet can be found in \autoref{tab:model}.

The primary goal of the AE module's training objective is to minimize the difference in reconstruction between the input and the reconstructed output. This study uses the mean square error (MSE) as the objective function for this module, which is given by: 
\begin{equation} \label{eq_mse}
\mathcal{L}_{\textrm{MSE}}(x,\hat{x}) = \frac{1}{N_{c}} \sum_{j=1}^{N_{c}} \| x_j - \hat{x_j} \|^2.
\end{equation}
Where $x_j$ denotes the input signals and $\hat{x_j}$ is the reconstructed signals of the channel $j$.

\begin{figure}[t]
\centering
\includegraphics[width=0.8\columnwidth]{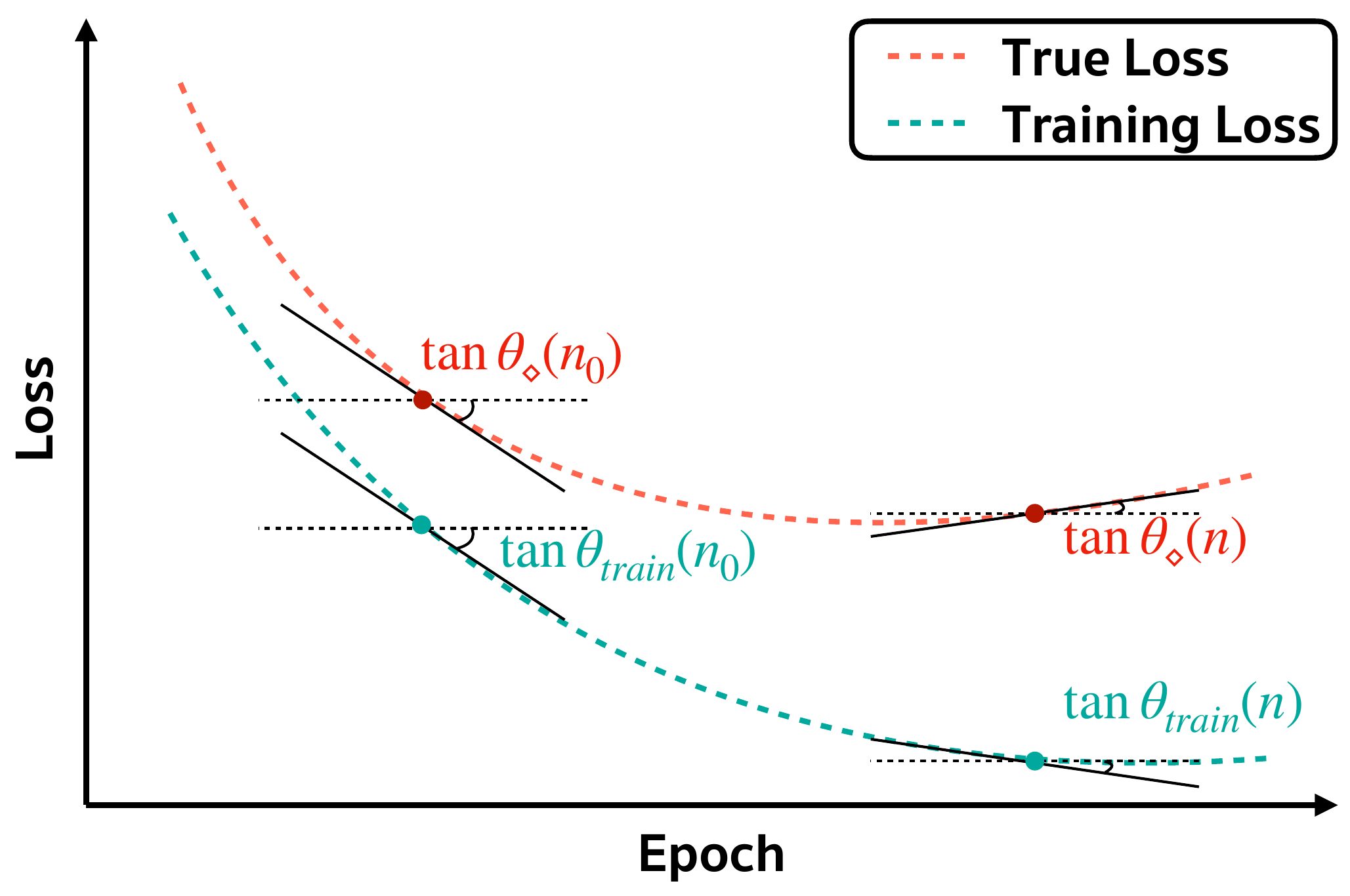}
\caption{The tangent slope of training and true losses at epoch $n$ and $n_0$.}
\label{fig:Generalization_overfitting_approximation}
\end{figure}

\subsubsection{Deep Metric Leaning}
Deep metric learning (DML) has proven its efficacy in retaining discriminative patterns for the latent representation of the autoencoder (AE), as indicated by our prior research\cite{min2net}. Hence, this research article employs a triplet loss approach, specifically using a semi-hard triplet constraint as proposed in the work of Schroff \textit{et al.}\cite{7298682}, inside the DML module. The aim is to maximize the relative distances between distinct classes of latent vectors. During the training procedure, a triplet set comprising three samples, represented as $\{x^a, x^p, x^n\}$, is randomly selected from the training dataset. The relationship between the anchor sample $x^a$ and the positive sample $x^p$ is ensured to be larger than the distance between the anchor sample $x^a$ and the negative sample $x^n$. Subsequently, the encoder component processes the set of three input signals concurrently, generating their corresponding latent vectors $z^a$, $z^p$, and $z^n$. The loss function can, therefore, be expressed as:

\begin{equation} \label{eq_triplet_loss}
\mathcal{L}_{\textrm{triplet}}(z^a,z^p,z^n) =  \frac{1}{2}\left[\|z^a - z^p\|^2 - \|z^a - z^n\|^2 + \alpha\right]_+.
\end{equation}
where $[z]_+ = max(z,0)$. The parameter $\alpha$ regulates as the margin threshold that imposes a constraint on the Euclidean distance $\|z^a - z^p\|^2$ between positive pairings, ensuring that it is less than the Euclidean distance $\|z^a - z^n\|^2$ between negative pairs. Determining the optimal value for the triplet loss margin is crucial to training the DML module.
\subsubsection{Supervised Learning} 
The supervised learning module employs a conventional soft-max classifier to classify the latent representations of the input spectral-spatial EEG signals. The latent representation $z$ is inputted into the fully connected (FC) layer using the softmax activation function to calculate the importance weight for each class. This can be represented as follows:
\begin{equation} \label{fc_softmax}
\hat{y}(z) = softmax(\mathcal{W}z+b)
\end{equation}
Where $\mathcal{W}$ and $b$ are the weight matrix and the bias vector, respectively. Then, the cross-entropy (CE) loss is used to measure the difference between the predicted value and true label, given by:
\begin{equation} \label{eq_cross-entropy}
\mathcal{L}_{\textrm{CE}}(y,\hat{y}) = -\frac{1}{N_{i}}\sum_{i=1}^{N_{i}} y_i\log{\hat{y}_i}.
\end{equation}
where $y$ represents the true label, $\hat{y}$ represents the classification probabilities, and $N_{i}$ represents the total number of input signals. The class with the highest classification probability is the predicted class of the single-trial EEG signal.

\subsection{Training Procedure for MixNet Using Adaptive Gradient Blending}
\label{sec:loss}
The training objective of the proposed method is optimized through the combination of the three loss functions: $\mathcal{L}_{\textrm{MSE}}$ in \autoref{eq_mse}, $\mathcal{L}_{\textrm{triplet}}$ in \autoref{eq_triplet_loss}, and $\mathcal{L}_{\textrm{CE}}$ in \autoref{eq_cross-entropy}. The final loss function of MixNet $\mathcal{L}_{\textrm{MixNet}}$ is represented as:
\begin{equation} \label{eq_total_loss}
\begin{aligned}
\mathcal{L}_{\textrm{MixNet}}(x,\hat{x},z^a,z^p,z^n,y,\hat{y}) =\frac{1}{N_{i}}\sum_{i=1}^{N_{i}}\{w^{(1)}\mathcal{L}_{\textrm{MSE}}(x_i,\hat{x_i})\\
+w^{(2)}\mathcal{L}_{\textrm{triplet}}(z_i^a,z_i^p,z_i^n)+w^{(3)} \mathcal{L}_{\textrm{CE}}(y_i,\hat{y_i})\}. 
\end{aligned}
\end{equation}
Where $N_{i}$ denotes the total number of input signals, $w^{(1)}$, $w^{(2)}$, and $w^{(3)}$ represent the hyperparameters to weigh the contribution of each loss function. As a result of integrating the three loss functions, both the unsupervised and the DML can influence the learning process when supervised learning occurs.

Manually tuning these weights $\{w^{(1)}, w^{(2)}, w^{(3)}\}$ is a problematic and expensive operation, making multi-tasking learning impractical. Inspired by the adaptive gradient blending concept, previously demonstrated in work by Huy \textit{et al.} \cite{adaptive_gradient_huy}, it has shown great success in regulating multiple loss weights simultaneously for joint multi-view learning. This study presents a systematic solution for multi-task learning, where the weighting of multiple loss functions is determined based on the adaptive gradient blending of each task. The concept of multi-task learning is designed to tackle the challenge of optimizing a model when faced with multiple objectives, which is a significant problem in many deep learning architectures. 

Acquiring access to the gradient flows of network streams becomes crucial for effectively regulating the learning pace. MixNet has three main learning tasks: autoencoder, deep metric learning, and supervised learning, optimizing three objective functions simultaneously. Conducting a thorough investigation of the tendencies towards generalization and overfitting in the learning tasks allows us to provide suitable weights to the gradient flows of these tasks. This means providing a higher weight to the task
that exhibits excellent generalization while a lower weight to the task with tendencies towards overfitting. In this approach, the gradients are blended in a way that considers the overall behavior of multi-task learning in terms of generalization and overfitting. Additionally, the learning progress of the network tasks is adjusted for each individual task.

This paper applies adaptive loss weighting to regulate the learning process across the three learning tasks to reach a balance in the generalization and overfitting rates inside the multi-task learning of MixNet. We determine the loss weights using a mechanism similar to that described in reference\cite{adaptive_gradient_huy} based on the ratio of generalization and overfitting metrics (a theoretical justification is provided in the supplementary material), as can be seen below:

\begin{equation} \label{eq_loss_weight_approximations}
\begin{aligned}
w^{(m)} = \frac{1}{Z}\frac{G_{m}}{O_{m}^{2}}
\end{aligned}
\end{equation}

where $m\in\{1,2,3\}$ and $Z$ is a normalization factor. The generalization metric is the gap between the gain in information learned about the target distribution. We define the overfitting metric as the gap between the gain on the training set and the target distribution. The weighted loss used through the training process at epoch $n$ is then expressed as follows:

\begin{equation} \label{eq_loss_weight_used_for_training}
\mathcal{L}(n) = \sum_{m\in\{1,2,3\}} w^{(m)}(n) \mathcal{L}^{(m)}(n)
\end{equation}

The learning behavior of MixNet can be influenced by the weighted loss mentioned above, relying on how $G_{m}$ and $O_{m}$ are approximated.

Generally, in the work \cite{wang2020makes}, $G_{m}$ and $O_{m}$ at the training process at epoch $n$ can be simply approximated as follows:

\begin{equation} \label{eq_Gm_approxomation_org}
\begin{aligned}
G_{m}(n) & \approx \mathcal{L}_{\diamond}^{(m)}(n_{0}) - \mathcal{L}_{\diamond}^{(m)}(n),
\end{aligned}
\end{equation}

\begin{equation} \label{eq_Om_approxomation_org}
\begin{aligned}
O_{m}(n) &\approx \Delta\mathcal{L}(n) - \Delta\mathcal{L}(n_{0}) \\ &\approx \left(\mathcal{L}_{\diamond}^{(m)}(n) - \mathcal{L}_{train}^{(m)}(n) \right) \\ &- \left(\mathcal{L}_{\diamond}^{(m)}(n_{0}) - \mathcal{L}_{train}^{(m)}(n_{0})\right)
\end{aligned}
\end{equation}

Where $\mathcal{L}_{train}^{(m)}$ and $\mathcal{L}_{\diamond}^{(m)}$ are defined as the training loss and true loss, respectively. Epoch $n$ is the training step and $n_{0}$ is reference step, where epoch $n_{0} < n$.
Nonetheless, the approximation above approach has two major drawbacks: the loss curves of training and validation are highly frustrated due to minibatch training, and the references $\mathcal{L}_{train}^{(m)}(0)$ and $\mathcal{L}_{\diamond}^{(m)}(0)$ may vary considerably with various random initializations. Alleviate these drawbacks; the powerful approximation approach is proposed to alleviate these drawbacks based on the tangents of losses over the line fitting of loss curves (training and validation) \cite{adaptive_gradient_huy}. This approximation method utilizes two fundamental observations: the smooth yet noisy pattern of loss curves and the informative characteristics of loss tangent directions at particular training iterations to determine the generalization or overfitting tendencies of the network. The approximation approach based on the tangents of losses throughout our experiments where $G_{m}$ and $O_{m}$ at the training process at epoch $n$ can be given by

\begin{equation} \label{eq_Gm_approxomation_tangent}
\begin{aligned}
G_{m}(n) & \approx \tan\theta_{\diamond}^{(m)}(n) - \tan\theta_{\diamond}^{(m)}(n_{0}),
\end{aligned}
\end{equation}

\begin{equation} \label{eq_Om_approxomation_tangent}
\begin{aligned}
O_{m}(n) &\approx \left( \tan\theta_{\diamond}^{(m)}(n) -  \tan\theta_{train}^{(m)}(n) \right) \\ &- \left( \tan\theta_{\diamond}^{(m)}(n_{0}) -  \tan\theta_{train}^{(m)}(n_{0}) \right)
\end{aligned}
\end{equation}

Where $\tan\theta_{train}^{(m)}(n)$ and $\tan\theta_{\diamond}^{(m)}(n)$ are the training and true loss tangents through the training process at epoch $n$ and  $\tan\theta_{train}^{(m)}(n_{0})$ and $\tan\theta_{\diamond}^{(m)}(n_{0})$ are their references, respectively. The relationship between all above variables can be summarized in \autoref{fig:Generalization_overfitting_approximation}. Moreover, let $\theta_{train}^{(m)}(n)$, and $\theta_{\diamond}^{(m)}(n)$, where $-90^{\circ}  \le \theta_{train}^{(m)}(n)  \le 0^{\circ}$ and $-90^{\circ}  \le \theta_{\diamond}^{(m)}(n)  \le 90^{\circ}$, denotes the angles made by the tangent lines of the training and true loss curves with the horizontal axis, respectively. The network is generalizing when $-90^{\circ}  \le \theta_{\diamond}^{(m)}(n)  < 0^{\circ}$ (i.e., negative tangent) and overfitting when $0^{\circ}  < \theta_{\diamond}^{(m)}(n)  \le 90^{\circ}$ (i.e., positive tangent).

\begin{algorithm}[t]
\caption{Process of Calculating MixNet's Loss Weights Using Adaptive Gradient Blending Approximation}\label{alg:1}
    \begin{algorithmic}[1]
    \small	
    \Procedure {MixNet\_Weight}{$\mathcal{L}_{train}, \mathcal{L}_{\diamond},W,l$}
    \State \textbf{input:} {$\mathcal{L}_{train}[1...n]$: array of training loss values 
    \State\hspace{3em}$\mathcal{L}_{\diamond}[1...n]$: array of true loss values} \State\hspace{3em}{$W$: warm-up period} 
    
    \State\hspace{3em}{$l$: window size for line fitting} 
    \State \textbf{output: }{$w(n)$: weight at the training epoch $n$}  
    \If {$n < W$} {$n_{0}=n$;}\hspace{1em}\Return 
    \EndIf
    \State $\tan\theta_{train}(n) \gets LineFit(\mathcal{L}_{train}[(n-l)...n])$
    \State $\tan\theta_{\diamond}(n) \gets LineFit(\mathcal{L}_{\diamond}[(n-l)...n])$
    \State\scriptsize  $G(n)=\tan\theta_{\diamond}(n) - \tan\theta_{\diamond}(n_{0})$
    \State\scriptsize $O(n)=[\tan\theta_{\diamond}(n)-\tan\theta_{train}(n)]-[\tan\theta_{\diamond}(n_{0})-\tan\theta_{train}(n_{0})]$
    \State $w(n) = \frac{1}{Z}\frac{G_{n}}{O_{n}}$
    \If {$\tan\theta_{\diamond}(n_{0}) > \tan\theta_{\diamond}(n)$} 
        \State $n_{0} = n$  \Comment{\footnotesize Training epoch w.r.t. the true loss's smallest tangent}
    \EndIf
    \EndProcedure
    \end{algorithmic}
\end{algorithm}

\begin{figure*}[ht]
\centering
\includegraphics[width=1.7\columnwidth]{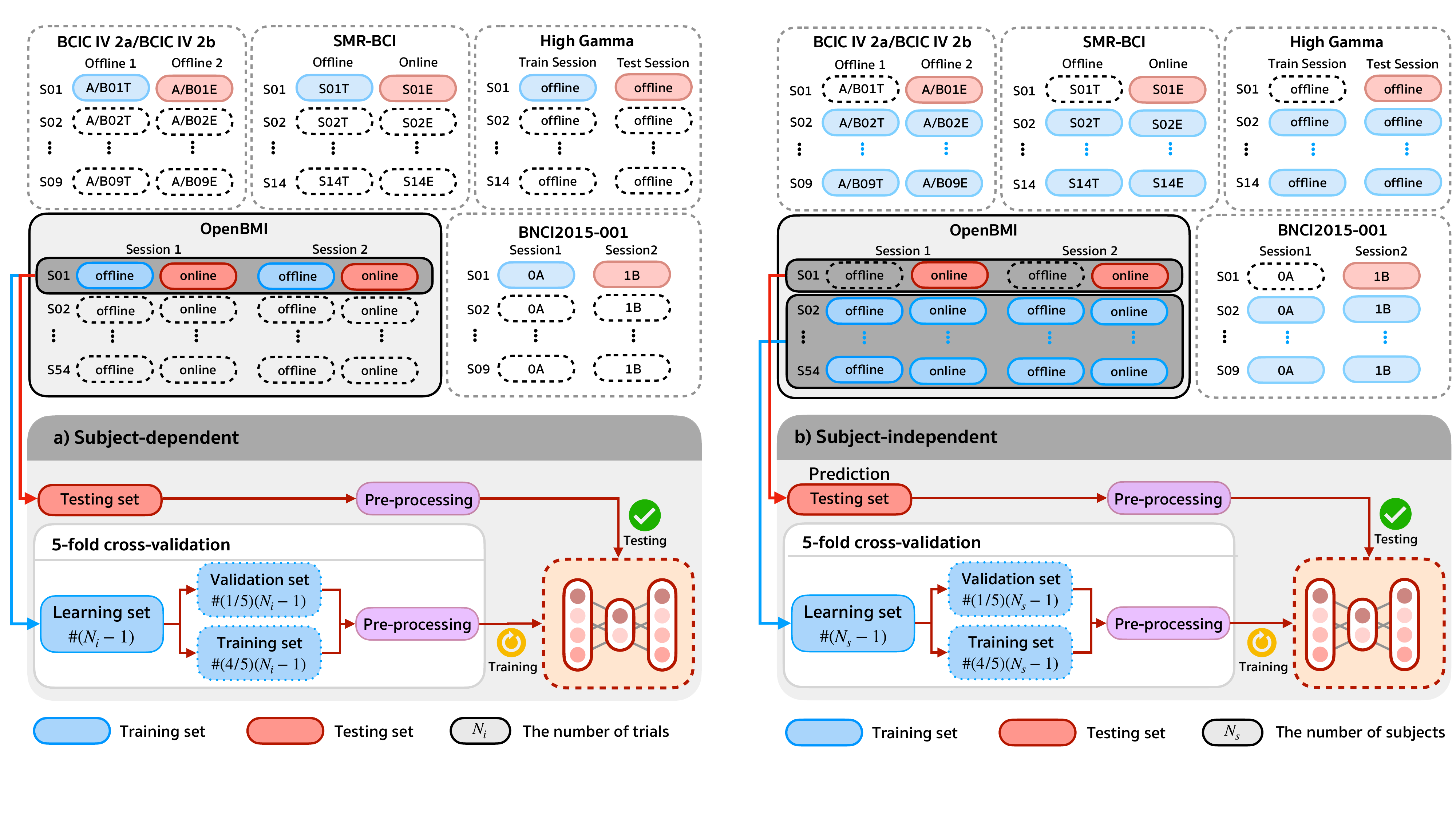}
\caption{Frameworks of a) subject-dependent and b) subject-independent with stratified k-fold cross-validation for the classification models. Note that the pre-processing procedure comprises both downsampling and FBCSP methods.}
\label{fig:k-fold}
\end{figure*}

The process of determining the adaptive loss weight, denoted as $w^{(m)}$, for a specific network task labeled as $m$ in MixNet, is detailed in Algorithm \ref{alg:1}. This involves estimating the loss tangent ($\tan\theta_{train}^{(m)}(n)$ and $\tan\theta_{\diamond}^{(m)}(n)$) at epoch $n$. This is done by analyzing the slope of lines fitted to a loss curve segment spanning length $l$ epoch, starting from epoch $n - l$ and extending to epoch $n$. This results in initiating the network's training with a warm-up period lasting $W$ training epochs, during which the task weights are uniformly set, where $l \le W$. Furthermore, for each learning task, the training epoch corresponding to the smallest authentic loss tangent (i.e., the steepest slope of the fitted line) is identified as the reference epoch $n_{0}$. In the practical context of MixNet, the true loss $\tan\theta_{\diamond}^{(m)}(n)$ remains unknown. Therefore, we approximate it with the loss measured on a held-out validation set $\tan\theta_{val}^{(m)}(n)$.

\subsection{Network Parameters}
The proposed method was implemented using TensorFlow framework version 2.8.2. An NVIDIA Tesla V100 GPU with 32GB of memory was utilized during the training process. We optimized the loss function in each training iteration using the Adam optimizer, with a learning rate schedule ranging from $10^{-3}$ to $10^{-4}$ throughout our experiments. The learning rate was reduced by a factor of 0.5 when no improvement in validation loss was observed for five consecutive epochs. A batch size of 32 samples was employed for the subject-dependent classification setting, while 100 samples were used for the subject-independent classification setting. The parameter $l$ in Algorithm \ref{alg:1}, which defines the window size for line fitting, was established at 1.5 epochs. Additionally, an $l$-point moving average was applied to the loss curves for smoothing purposes prior to conducting line fitting. Finally, we employed the early stopping strategy to determine the number of training iterations. The training process was halted if there was no reduction in validation loss for 20 consecutive epochs.

\subsection{Baseline Approaches}
To validate the effectiveness of MixNet, we implemented five state-of-the-art methods for performance comparison. 

\subsubsection{FBCSP-SVM}
FBCSP builds on the original CSP algorithm \cite{4634130}, extracting discriminative EEG features from multiple frequency bands. We implemented FBCSP using MNE-Python package (version 0.20) \cite{mne_package}, applying four spatial filters to decompose EEG signals into nine 4 Hz-wide bands (4--8 Hz, 8--12 Hz, ..., 36--40 Hz) via 5\textsuperscript{th} order non-causal Butterworth bandpass filtering. Furthermore, a support vector machine (SVM) with a grid search considered kernel options (linear, radial bias function (RBF), sigmoid), $C$ values (0.001, 0.01, 0.1, 1, 10, 100, 1000), and, particularly for RBF, gamma values (0.01, 0.001). The grid search optimized hyperparameters through validation set predictions. The resulting SVM classifier, with optimal parameters, was then used for testing.

\subsubsection{DeepConvNet}
The DeepConvNet, introduced initially as a deep learning model based on two CNN architectures\cite{tonio_paper}, has demonstrated considerable efficacy in EEG-MI classification. In this current study, we faithfully employed the DeepConvNet with optimal parameters as outlined in \cite{tonio_paper}. Moreover, to prepare the raw EEG data for analysis, we applied a bandpass filter spanning the frequency range of 8 to 30 Hz, utilizing a 5\textsuperscript{th} order non-causal Butterworth filter.

\subsubsection{EEGNet-8,2}
Inspired by the FBCSP method, EEGNet-8,2 was introduced as a compact CNN architecture to capture distinctive EEG features, yielding remarkable performance across various BCI paradigms \cite{Lawhern_2018}. In this context, we reproduced the EEGNet-8,2 architecture to achieve comparable results. The network's hyperparameters were maintained in the optimal configuration recommended in the original publication \cite{Lawhern_2018}. Moreover, the raw EEG data performed similar pre-processing to the DeepConvNet model, ensuring consistent input preparation for the EEGNet-8,2 training process.

\subsubsection{Spectral-spatial CNN}
The spectral-spatial CNN framework, built upon CNN architectures, was introduced by \cite{8897723}, demonstrating outstanding performance in subject-independent MI decoding. This framework learned the spectral-spatial input, capturing discriminative features from EEG signals across multiple frequency bands. This paper transformed the raw EEG signals into a similar spectral-spatial representation as conducted in \cite{8897723}. The spectral-spatial CNN model was implemented using the optimal parameters defined in the original paper.

\subsubsection{MIN2Net}
MIN2Net, introduced in our previous research\cite{min2net}, was developed as an advanced end-to-end neural network architecture based on a multi-task autoencoder framework. Within this architecture, MIN2Net concurrently accomplishes unsupervised EEG-MI reconstruction and supervised EEG-MI classification tasks, leading to the capture of meaningful features. The demonstrated performance of MIN2Net is excellent in extracting discriminative features for subject-independent MI decoding.

\subsection{Performance Evaluation}
To validate MixNet's performance in MI classification, we conducted experiments using subject-dependent and subject-independent approaches on six benchmark datasets: BCIC IV 2a, BCIC IV 2b, BNCI2015-001, SMR-BCI, High-Gamma, and OpenBMI. \autoref{fig:k-fold} illustrates how we divided the training and testing sets for both manners.

The training and testing sets were sourced from the same subject but in different sessions in the subject-dependent setup. For subject-independent evaluation, we utilized leave-one-subject-out cross-validation (LOSO-CV). In both scenarios, we applied stratified 5-fold cross-validation (CV) to partition the training set into new training and validation subsets, ensuring balanced class representation. Lastly, we evaluated all the considered methods' performance using Accuracy, F1-score, and area under the curve (AUC), offering a comprehensive assessment of their effectiveness\cite{f1_ref}.

This paper consists of three main experiments as follows:
\subsubsection{Experiment I}
In pursuit of optimal hyperparameter settings for MixNet, we conducted an ablation study to refine parameter choices. Initially, we explored the influence of the number of spatial filters, denoted as $U$ in \autoref{last_csp}. A grid search was conducted over $\{2, 4, 6, 8, 10\}$ to determine the optimal value of $U$.

As evident from \autoref{eq_triplet_loss}, the margin parameter $\alpha$ significantly impacts MixNet's training. We systematically investigated $\alpha$ via experiments in MixNet, seeking the optimal value from the set $\{0.1, 0.5, 1.0, 5.0, 10.0, 100.0\}$.

The latent vector size $z$ within MixNet's autoencoder module is vital in training, preserving essential features in high-dimensional EEG-MI signals. We performed a grid search over $\{8, U\times{N_{b}}, 64, 128, 256\}$ to ascertain an optimal latent vector size. The optimal latent vector size aims to reduce dimensions while maintaining the quality of data representation.

The parameter $W$, representing the warm-up period, was integral to Algorithm \ref{alg:1}. We conducted a grid search over the set $\{2, 3, 5, 7, 9\}$ to determine the optimal $W$ size.

\subsubsection{Experiment II}
A comparative analysis was conducted to evaluate the classification performance of MixNet and all baseline methods. The study centered on binary MI classification using high-density EEG recordings. It aimed to investigate the effectiveness of all the approaches considered across five benchmark datasets (BCIC IV 2a, BNCI2015-001, SMR-BCI, High-Gamma, and OpenBMI), covering subject-dependent and subject-independent scenarios. We assessed all methods using identical training, validation, and testing sets to ensure a fair comparison.

\subsubsection{Experiment III}
Low-density EEG data collection needs reduced sensors or electrodes, resulting in less expensive setup and equipment. This makes deploying EEG systems in various real-world scenarios, such as home monitoring, remote healthcare, and mobile applications, more feasible and cost-effective\cite{venkatesh2022ssvep, he2023diversity}. However, the low-density EEG system might sacrifice some spatial resolution, resulting in suboptimal classification performance. 

To demonstrate the practicality of MixNet in developing real-world applications, we conducted a binary classification task on three-channel EEG-MI in both subject-dependent and subject-independent manners. The objective of this experiment was to evaluate the effectiveness of MixNet compared to all other baseline methods in a low-density EEG system scenario. By considering the three-channel EEG-MI dataset, it is important to note that the number of spatial filters $U$ was set to 2. Additionally, the optimal MI classification performance of MixNet across all hyperparameter searches ($\alpha$, $z$, $W$) for a three-channel EEG-MI dataset is reported in the supplementary materials.

\subsection{Visualization}
To compare the efficacy of deep learning techniques in extracting highly discriminative features from EEG signals, we used the $t$-SNE method\cite{vandermaaten08a} to visually represent the generalized brain features learned by different deep learning methods within a two-dimensional embedding space. Across all trained models, the $t$-SNE algorithm was used to visualize the high-dimensional embedding space at the input of the FC layer.
\begin{table}[t]
\caption{Classification performance (Accuracy $\pm$ SD and F1-score $\pm$ SD) in \% of MixNet using the subject-dependent and subject-independent manners comparisons on five different CSP components ($U$). Bold denotes the best numerical values.}
    \label{tab:adaptive_min2net_vary_csp_components}
    \centering
    \Large
    \resizebox{1.0\columnwidth}{!}{
\begin{tabular}{@{}cccccc@{}}
\toprule[0.2em]
\multirow{2}{*}{\textbf{Datasets}} & \multirow{2}{*}{\textbf{\# of component}} & \multicolumn{2}{c}{\textbf{Subject-dependent}}  & \multicolumn{2}{c}{\textbf{Subject-independent}} \\ \cmidrule[0.1em](l){3-4} \cmidrule[0.1em](l){5-6}
                                   &                                           & \textbf{Accuracy}      & \textbf{F1-score}      & \textbf{Accuracy}       & \textbf{F1-score}      \\ \midrule[0.1em]
\multirow{5}{*}{BCIC IV 2a}            & \textbf{2}                                         & \textbf{77.35 ± 14.41} & \textbf{76.36 ± 15.62} & 67.11 ± 11.46          & 64.31 ± 14.19          \\ 
                                   & \textbf{4}                                         & 76.16 ± 12.81          & 75.30 ± 13.54          & \textbf{69.07 ± 12.10} & \textbf{67.24 ± 14.38} \\ 
                                   & 6                                         & 76.08 ± 14.01          & 75.03 ± 15.24          & 67.93 ± 12.73          & 65.13 ± 15.77          \\ 
                                   & 8                                         & 75.71 ± 13.24          & 74.43 ± 14.64          & 68.30 ± 12.75          & 66.01 ± 15.05          \\ 
                                   & 10                                        & 74.81 ± 13.15          & 73.55 ± 14.53          & 68.38 ± 11.72          & 66.05 ± 14.44          \\ \midrule[0.1em]
                                   
\multirow{5}{*}{BNCI2015-001}      & \textbf{2}            & 75.66 ± 15.85  & 73.59 ± 18.51          & \textbf{64.67 ± 10.81}          & \textbf{60.46 ± 14.60}          \\ 
                                   & \textbf{4}            & \textbf{78.38 ± 14.18}  & \textbf{76.90 ± 16.29}          & 63.78 ± 10.20          & 59.69 ± 13.63          \\ 
                                   & 6                     & 77.88 ± 15.94  & 76.20 ± 18.41          & 62.34 ± 9.79           & 57.96 ± 13.00          \\ 
                                   & 8                     & 77.46 ± 15.16  & 75.91 ± 17.17          & 61.39 ± 9.71           & 57.06 ± 12.96          \\ 
                                   & 10                    & 75.78 ± 15.77  & 73.77 ± 18.59          & 61.39 ± 9.23           & 56.71 ± 12.94          \\ \midrule[0.1em]
                                   
\multirow{5}{*}{SMR-BCI}          & \textbf{2}                                         & \textbf{75.24 ± 16.45} & \textbf{73.76 ± 18.33} & 61.02 ± 14.30          & 55.67 ± 18.13           \\ 
                                   & 4                                         & 74.69 ± 17.49          & 73.52 ± 18.83          & 65.17 ± 14.01          & 60.89 ± 17.82           \\ 
                                   & \textbf{6}                                         & 72.00 ± 16.73          & 70.42 ± 18.33          & \textbf{66.02 ± 14.69} & \textbf{61.56 ± 18.72}  \\ 
                                   & 8                                         & 71.62 ± 14.90          & 70.31 ± 15.99          & 64.81 ± 13.62          & 59.05 ± 18.97           \\ 
                                   & 10                                        & 69.26 ± 16.22          & 67.58 ± 17.83          & 63.79 ± 14.49          & 57.37 ± 19.97           \\ \midrule[0.1em]
\multirow{5}{*}{High-Gamma}         & 2                                         & 77.64 ± 16.71          & 77.05 ± 17.33          & 65.32 ± 12.62          & 62.50 ± 15.31          \\ 
                                   & 4                                         & 79.45 ± 16.14          & 79.05 ± 16.54          & 66.86 ± 11.02	          & 64.75 ± 13.00          \\ 
                                   & \textbf{6}                                         & \textbf{80.23 ± 15.28} & \textbf{79.87 ± 15.67} & 69.04 ± 12.06          & 67.14 ± 14.02          \\ 
                                   & 8                                         & 79.66 ± 15.87          & 79.00 ± 16.76          & 69.57 ± 11.62       & 67.92 ± 13.52 \\ 
                                   & \textbf{10}                                        & 78.82 ± 16.31          & 78.24 ± 16.91          & \textbf{69.84 ± 10.82}   & \textbf{67.97 ± 12.98}          \\ \midrule[0.1em]
\multirow{5}{*}{OpenBMI}           & \textbf{2}                                         & 67.94 ± 16.98          & 67.31 ± 17.40          & \textbf{71.09 ± 14.29} & \textbf{70.12 ± 15.23} \\ 
                                   & \textbf{4}                                         & \textbf{68.56 ± 16.70} & \textbf{67.73 ± 17.37} & 70.41 ± 13.82          & 69.78 ± 14.34          \\ 
                                   & 6                                         & 68.07 ± 16.78          & 67.25 ± 17.43          & 69.45 ± 13.48          & 68.89 ± 13.88          \\ 
                                   & 8                                         & 67.47 ± 16.74          & 66.40 ± 17.60          & 69.62 ± 13.69          & 69.04 ± 14.15          \\ 
                                   & 10                                        & 66.78 ± 16.61          & 65.82 ± 17.32          & 70.14 ± 13.78          & 69.54 ± 14.28          \\ \bottomrule[0.2em]
\end{tabular}}
\end{table}

\begin{table}[t]
\caption{Classification performance (Accuracy $\pm$ SD and F1-score $\pm$ SD) in \% of MixNet using the subject-dependent and subject-independent manners comparisons on six different margins ($\alpha$). Bold denotes the best numerical values.}
    \label{tab:AE_triplet_margin_results}
    \centering
    \Large
    \resizebox{1.0\columnwidth}{!}{
\begin{tabular}{@{}cccccc@{}}
\toprule[0.2em]
\multirow{2}{*}{\textbf{Datasets}} & \multirow{2}{*}{\textbf{Margins}} & \multicolumn{2}{c}{\textbf{Subject-dependent}}  & \multicolumn{2}{c}{\textbf{Subject-independent}} \\ \cmidrule[0.1em](l){3-4} \cmidrule[0.1em](l){5-6} 
                                   &                                   & \textbf{Accuracy}      & \textbf{F1-score}      & \textbf{Accuracy}       & \textbf{F1-score}      \\ \midrule[0.1em]
\multirow{6}{*}{BCIC IV 2a}            & \textbf{0.1}                               & 76.48 ± 15.17          & 75.48 ± 16.38           & \textbf{69.23 ± 11.80} & \textbf{67.42 ± 13.88} \\ 
                                   & 0.5                               & 77.30 ± 15.22          & 76.32 ± 16.37           & 68.09 ± 12.08          & 65.79 ± 14.52          \\ 
                                   & \textbf{1}                                 & \textbf{77.35 ± 14.41} & \textbf{76.36 ± 15.62}  & 69.07 ± 12.10          & 67.24 ± 14.38          \\ 
                                   & 5                                 & 75.65 ± 14.53          & 74.36 ± 16.08           & 67.69 ± 11.90          & 65.81 ± 13.90          \\ 
                                   & 10                                & 76.40 ± 14.50          & 75.30 ± 15.72           & 67.55 ± 11.56          & 65.61 ± 13.57          \\ 
                                   & 100                               & 75.23 ± 14.40          & 74.43 ± 15.15           & 67.85 ± 12.70          & 65.87 ± 14.74          \\ \midrule[0.1em]
                                   
\multirow{6}{*}{BNCI2015-001}      & 0.1                               & 77.08 ± 14.48          & 75.58 ± 16.16          & 64.67 ± 11.45          & 60.44 ± 15.08 \\ 
                                   & 0.5                               & 78.10 ± 15.00          & 76.31 ± 18.26          & 64.76 ± 10.11          & 60.88 ± 13.27          \\ 
                                   & 1                                 & 78.38 ± 14.18          & 76.90 ± 16.29           & 64.67 ± 10.81          & 60.46 ± 14.60          \\ 
                                   & 5                                 & 79.00 ± 13.27          & 77.73 ± 15.36           & 65.64 ± 11.25          & 61.46 ± 15.04          \\ 
                                    & \textbf{10}                      & \textbf{79.03 ± 13.29} & \textbf{77.97 ± 14.93}           & 64.98 ± 10.68          & 61.15 ± 13.84          \\ 
                                   & \textbf{100}                               & 76.20 ± 13.86          & 74.56 ± 16.04           & \textbf{65.85 ± 11.23}          & \textbf{62.08 ± 14.64}          \\ \midrule[0.1em]
                                   
\multirow{6}{*}{SMR-BCI}          & 0.1                               & 75.45 ± 16.49          & 73.84 ± 18.58           & 65.93 ± 15.88          & 60.80 ± 20.71          \\ 
                                   & \textbf{0.5}                               & \textbf{75.40 ± 16.33} & \textbf{74.14 ± 17.95}  & 64.98 ± 14.99          & 60.09 ± 19.53          \\ 
                                   & 1                                 & 75.24 ± 16.45          & 73.76 ± 18.33           & 66.02 ± 14.69          & 61.56 ± 18.72          \\ 
                                   & \textbf{5}                                 & 74.38 ± 16.94          & 72.65 ± 19.14           & \textbf{66.14 ± 14.00} & \textbf{62.14 ± 17.73} \\ 
                                   & 10                                & 73.36 ± 16.58          & 71.74 ± 18.43           & 65.31 ± 13.97          & 61.06 ± 17.88          \\ 
                                   & 100                               & 74.88 ± 16.83          & 73.32 ± 18.83           & 63.36 ± 13.93          & 58.06 ± 18.18          \\ \midrule[0.1em]
\multirow{6}{*}{High-Gamma}         & 0.1                               & 79.95 ± 15.73          & 79.59 ± 16.12           & 68.73 ± 11.73         & 66.41 ± 14.41 \\ 
                                   & 0.5                               & 79.61 ± 15.80          & 79.02 ± 16.59           & 69.21 ± 10.54	        & 67.04 ± 13.22          \\ 
                                   & \textbf{1}                                 & \textbf{80.23 ± 15.28} & \textbf{79.87 ± 15.67}  & \textbf{69.84 ± 10.82}	          & \textbf{67.97 ± 12.98}          \\
                                   & 5                                 & 79.66 ± 16.56          & 79.16 ± 17.07           & 68.98 ± 11.51          & 66.77 ± 14.00          \\ 
                                   & 10                                & 79.84 ± 16.09          & 79.46 ± 16.52           & 69.57 ± 11.08          & 67.85 ± 12.89          \\ 
                                   & 100                               & 77.96 ± 16.87          & 77.57 ± 17.25           & 69.41 ± 11.90          & 66.91 ± 14.75          \\ \midrule[0.1em]
\multirow{6}{*}{OpenBMI}           & 0.1                               & 68.47 ± 16.57          & 67.83 ± 17.06          & 70.59 ± 14.11          & 69.65 ± 14.97          \\ 
                                   & 0.5                               & 68.57 ± 16.36          & 67.86 ± 16.91          & 70.83 ± 14.00          & 69.97 ± 14.76          \\ 
                                   & 1                                 & 68.56 ± 16.70          & 67.73 ± 17.37          & 71.09 ± 14.29          & 70.12 ± 15.23          \\ 
                                   & \textbf{5}                                 & \textbf{68.95 ± 16.75} & \textbf{68.04 ± 17.54} & 71.25 ± 14.33          & 70.32 ± 15.21          \\ 
                                   & 10                                & 68.70 ± 16.69          & 67.84 ± 17.36          & 71.25 ± 14.33          & 70.36 ± 15.14          \\ 
                                   & \textbf{100}                               & 68.17 ± 16.21          & 67.12 ± 17.12          & \textbf{71.95 ± 14.18} & \textbf{71.13 ± 14.88} \\ \bottomrule[0.2em]
\end{tabular}}
\end{table}

\section{Results}
This section presents the findings and statistical analysis of Experiments I, II, and III, which aim to establish the efficacy of MixNet. Moreover, we employ visualization techniques to illustrate the discriminative pattern of the EEG features acquired by the MixNet model compared to other baseline methods. The results of each experiment were reported in terms of accuracy, F1-score, and AUC, along with their respective standard deviations (Accuracy $\pm$ SD, F1-score $\pm$ SD, AUC $\pm$ SD).

\subsection{Experiment I: Ablation Study}
\autoref{tab:adaptive_min2net_vary_csp_components} presents the classification results for varying values of the number of spatial filters, denoted as $U$, incorporated into the spectral-spatial signals of MixNet. The impact of different values of spatial filters on the final classification performance is evident. Notably, the choice of $U$ values significantly influenced the outcomes. MixNet demonstrated its highest performance in a subject-dependent manner by utilizing $U=2$ for the BCIC IV 2a and SMR-BCI datasets, $U=4$ for the BNCI2015-001 and OpenBMI datasets, and $U=6$ for the High-Gamma dataset. In a subject-independent scenario, MixNet achieved optimal performance with $U=2$ on the BNCI2015-001 and OpenBMI datasets, while on the BCIC IV 2a dataset, $U=4$ yielded the best results. For the SMR-BCI and High-Gamma datasets, the most effective $U$ values were 6 and 10, respectively, contributing to MixNet's superior performance.

\autoref{tab:AE_triplet_margin_results} depicts the classification results obtained when utilizing different values of the margin parameter ($\alpha$) in the DML module of MixNet. We found that the value of the margin parameter significantly impacts the classification performance. Utilizing a margin value of 1.0 for the BCIC IV 2a and High-Gamma datasets, 0.5 for the SMR-BCI dataset, 5 for the OpenBMI dataset, and a margin value of 10 for the BNCI2015-001 dataset, MixNet demonstrated its optimal subject-dependent performance. Furthermore, MixNet demonstrated its most effective subject-independent performance by utilizing margin values of 0.1 for the BCIC IV 2a dataset, 1.0 for the High-Gamma dataset, 5 for the SMR-BCI dataset, and 100 for the BNCI2015-001 and OpenBMI datasets.

\autoref{tab:AE_triplet_latent_dims_results} presents the averaged classification performance corresponding to different latent vector $z$ sizes within MixNet's autoencoder module. It can be observed that the latent vector's dimensions significantly impact the averaged classification performance. Notably, when the latent vector size is defined as $z = U \times N_{b}$, MixNet achieves its optimal performance in a subject-dependent manner across all the considered datasets. MixNet achieves its highest performance in a subject-independent manner by configuring $z = U \times N_{b}$ for the SMR-BCI, High-Gamma, and OpenBMI datasets. Conversely, the best MixNet performance on the BCIC IV 2a and BNCI2015-001 datasets is reached when $z = 128$ is employed.

\autoref{tab:MixNet_warmup_results} presents the average classification performance across different warm-up period sizes ($W$) within adaptive gradient blending. The size of the warm-up period significantly influences the average classification performance. Optimal subject-dependent performance was observed with a warm-up size of 2 for the BNCI2015-001 and SMR-BCI datasets, 5 for the High-Gamma and OpenBMI datasets, and 7 for the BCIC IV 2a dataset. Remarkably, MixNet achieves its optimal subject-independent performance with a warm-up period size of 5 across all evaluated datasets.

\begin{table}[t]
\caption{Classification performance (Accuracy $\pm$ SD and F1-score $\pm$ SD) in \% of MixNet using the subject-dependent and subject-independent manners comparisons on five different sizes of latent vector ($z$). Bold denotes the best numerical values.}
    \label{tab:AE_triplet_latent_dims_results}
    \centering
    \Large
    \resizebox{1.0\columnwidth}{!}{
\begin{tabular}{@{}cccccc@{}}
\toprule[0.2em]
\multirow{2}{*}{\textbf{Datatsets}} & \multirow{2}{*}{\textit{\textbf{\begin{tabular}[c]{@{}c@{}}\# of latent\\ vector\end{tabular}}}} & \multicolumn{2}{c}{\textbf{Subject-dependent}} & \multicolumn{2}{c}{\textbf{Subject-independent}} \\ \cmidrule[0.1em](l){3-4} \cmidrule[0.1em](l){5-6} 
                                    &                                                & \textbf{Accuracy}      & \textbf{F1-score}     & \textbf{Accuracy}       & \textbf{F1-score}      \\ \midrule[0.1em]
\multirow{5}{*}{BCIC IV 2a}             & 8                                              & 76.93 ± 15.03          & 75.74 ± 16.45         & 68.64 ± 12.27           & 66.40 ± 14.80          \\ 
                                    & \textbf{$U\times N_f$}                                             & \textbf{77.35 ± 14.41} & \textbf{76.36 ± 15.62} & 69.23 ± 11.80          & 67.42 ± 13.88          \\ 
                                    & 64                                             & 76.60 ± 14.95          & 75.63 ± 16.19         & 68.43 ± 12.32           & 65.89 ± 15.40          \\ 
                                    & \textbf{128}                                             & 76.62 ± 14.26          & 75.64 ± 15.38         & \textbf{69.35 ± 11.79}  & \textbf{67.54 ± 13.86} \\ 
                                    & 256                                            & 76.37 ± 14.58          & 75.58 ± 15.43         & 68.92 ± 11.27           & 67.08 ± 13.46          \\ \midrule[0.1em]
                                    
\multirow{5}{*}{BNCI2015-001}        & 8   & 78.50 ± 14.03          & 77.06 ± 16.03                     & 64.53 ± 11.52           & 59.93 ± 15.30          \\ 
                                    & \textbf{$U\times N_f$}                                             & \textbf{79.03 ± 13.29} & \textbf{77.97 ± 14.93} & 65.85 ± 11.23          & 62.08 ± 14.64          \\ 
                                    & 64                                             & 79.12 ± 13.78          & 77.64 ± 15.88         & 65.47 ± 11.04           & 61.64 ± 14.50          \\ 
                                    & \textbf{128}                                             & 77.88 ± 14.55          & 76.03 ± 17.59         & \textbf{66.22 ± 11.70}  & \textbf{62.55 ± 14.95} \\ 
                                    & 256                                            & 77.25 ± 14.89          & 75.18 ± 18.08         & 64.55 ± 11.90           & 60.49 ± 15.57          \\ \midrule[0.1em]
                                    
\multirow{5}{*}{SMR-BCI}           & 8                                              & 75.62 ± 16.11          & 73.86 ± 18.33         & 63.90 ± 13.77           & 59.02 ± 18.35          \\ 
                                    & \textbf{$U\times N_f$}                                             & \textbf{75.40 ± 16.33} & \textbf{74.14 ± 17.95} & \textbf{66.14 ± 14.00} & \textbf{62.14 ± 17.73} \\ 
                                    & 64                                             & 75.14 ± 16.32          & 73.81 ± 18.02         & 66.17 ± 14.17           & 62.06 ± 18.31          \\ 
                                    & 128                                            & 74.17 ± 16.52          & 72.95 ± 18.02         & 64.52 ± 13.22           & 59.76 ± 17.36          \\ 
                                    & 256                                            & 73.71 ± 16.88          & 72.52 ± 18.33         & 64.26 ± 12.97           & 59.69 ± 17.01          \\ \midrule[0.1em]
\multirow{5}{*}{High-Gamma}          & 8                                              & 79.57 ± 15.77          & 79.02 ± 16.38         & 68.91 ± 11.15           & 66.93 ± 13.42          \\ 
                                    & \textbf{$U\times N_f$}                                             & \textbf{80.23 ± 15.28}          & \textbf{79.87 ± 15.67}         & \textbf{69.84 ± 10.82}  & \textbf{67.97 ± 12.98} \\ 
                                    & 64                                             & 79.79 ± 15.55 & 79.19 ± 16.48 & 69.18 ± 11.51          & 67.31 ± 13.62          \\ 
                                    & 128                                            & 78.45 ± 16.14	          & 77.82 ± 16.76         & 69.18 ± 11.58           & 67.02 ± 14.13          \\ 
                                    & 256                                            & 79.21 ± 16.19          & 78.60 ± 17.08         & 69.55 ± 10.78           & 67.70 ± 12.79          \\ \midrule[0.1em]
\multirow{5}{*}{OpenBMI}            & 8                                              & 68.52 ± 16.54          & 67.73 ± 17.12         & 71.62 ± 14.25           & 70.75 ± 15.04          \\ 
                                    & \textbf{$U\times N_f$}                                             & \textbf{68.95 ± 16.75} & \textbf{68.04 ± 17.54} & \textbf{71.95 ± 14.18} & \textbf{71.13 ± 14.88} \\ 
                                    & 64                                             & 68.40 ± 16.98          & 67.67 ± 17.56         & 71.82 ± 14.47           & 70.88 ± 15.38          \\ 
                                    & 128                                            & 68.69 ± 16.81          & 67.86 ± 17.49         & 71.75 ± 14.27           & 70.91 ± 15.00          \\ 
                                    & 256                                            & 68.77 ± 16.97          & 68.04 ± 17.55         & 71.68 ± 14.49           & 70.66 ± 15.49          \\ \bottomrule[0.2em]
\end{tabular}}
\end{table}

\subsection{Experiment II: Binary MI Classification}

The overall classification performance of MixNet and five baseline methods, using high-density EEG-MI across all five datasets in both subject-dependent and subject-independent settings, is presented in \autoref{tab:result_binary_classification_20chs}. In subject-dependent analyses, MixNet exhibited superior accuracy and F1-score across the BCIC IV 2a, BNCI2015-001, High-Gamma, and OpenBMI datasets. Remarkably, MixNet achieved its best performance on the SMR-BCI dataset in terms of F1-score. Furthermore, MixNet's performance in terms of AUC outperformed that of other baseline methods on the High-Gamma and OpenBMI datasets.


The subject-independent analysis results indicate that MixNet displayed superior performance in accuracy, F1-score, and AUC on the BCIC IV 2a and High-Gamma datasets. For the BNCI2015-001 dataset, MixNet achieved its highest performance in accuracy and F1-score, and the second-highest in AUC. In evaluating the SMR-BCI dataset, MixNet surpassed all baseline methods in F1-score, while it secured the second-highest performance in accuracy and AUC. Additionally, on the OpenBMI dataset, MixNet outperformed all baseline methods in accuracy, F1-score, and AUC, with the exception of the MIN2Net approach.

\begin{figure}[t]
\centering
\includegraphics[width=1.00\columnwidth]{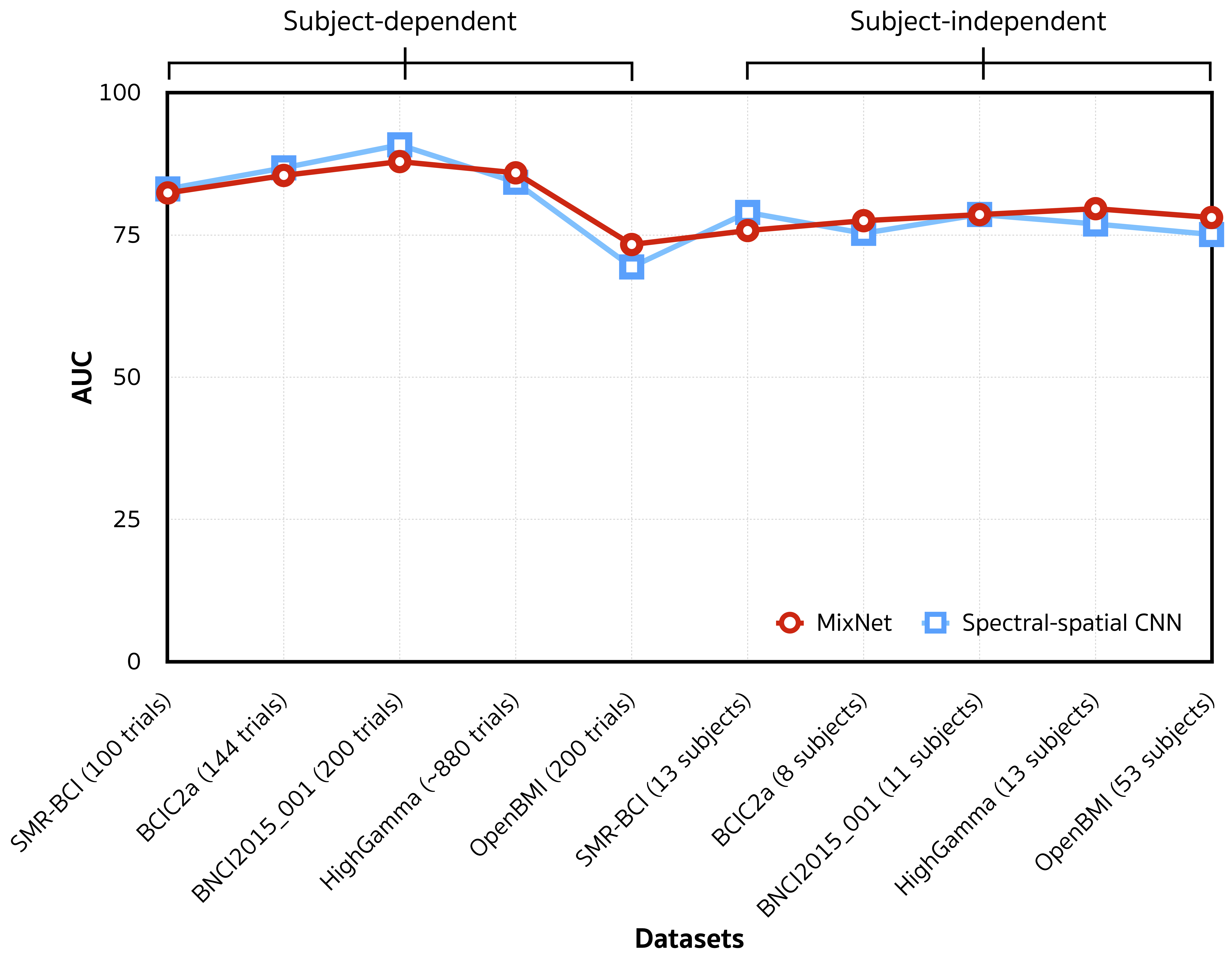}
\caption{Impact of the number of training samples on binary classification performance, measured by AUC score, across two considered methods.}
\label{fig:Line_chart_comparing_MixNet_SpectralSpatialCNN_AUC_metric}
\end{figure}

\autoref{fig:Line_chart_comparing_MixNet_SpectralSpatialCNN_AUC_metric} illustrates the variation in classification performance, measured by the AUC metric, relative to the number of training samples. The $x$-axis denotes the number of training samples across all datasets, while the $y$-axis represents the binary classification AUC for MixNet and Spectral-Spatial CNN methods. It can be seen that MixNet consistently outperforms the Spectral-Spatial CNN method in terms of AUC in both subject-dependent and subject-independent scenarios, particularly with larger datasets. Therefore, the volume of training samples is a key factor in enhancing MixNet's classification efficacy, demonstrating its robustness with extensive training data.

\subsection{Experiment III: Low-density EEG-MI Classification}
\autoref{tab:result_binary_classification_3chs} illustrates the binary classification performance of MixNet and five baseline methods using low-density EEG (3-channel) across the BCIC IV 2b dataset. This evaluation covers both subject-dependent and subject-independent scenarios. It can be seen that MixNet outperforms all baseline methods in terms of accuracy and F1-score in the subject-dependent setting. In the subject-independent manner, MixNet achieved the highest performance in terms of accuracy, F1-score, and AUC. This result indicates the promising potential of MixNet in developing real-world applications.

\begin{table}[t]
\caption{Classification performance (Accuracy $\pm$ SD and F1-score $\pm$ SD) in \% of MixNet using the subject-dependent and subject-independent manners comparisons on five different sizes of warm-up period ($W$). Bold denotes the best numerical values.}
    \label{tab:MixNet_warmup_results}
    \centering
    \Large
    \resizebox{1.0\columnwidth}{!}{
\begin{tabular}{@{}cccccc@{}}
\toprule[0.2em]
\multirow{2}{*}{\textbf{Datatsets}} & \multirow{2}{*}{\textbf{Warm-up}} & \multicolumn{2}{c}{\textbf{Subject-dependent}}  & \multicolumn{2}{c}{\textbf{Subject-independent}} \\ \cmidrule[0.1em](l){3-4} \cmidrule[0.1em](l){5-6} 
                                    &                                   & \textbf{Accuracy}      & \textbf{F1-score}      & \textbf{Accuracy}       & \textbf{F1-score}      \\ \midrule[0.1em]
\multirow{5}{*}{BCIC IV 2a}             & 2                                 & 76.20 ± 15.21          & 74.99 ± 16.65          & 69.23 ± 11.97           & 67.47 ± 13.95          \\ 
                                    & 3                                 & 77.07 ± 14.60          & 76.17 ± 15.77          & 68.94 ± 11.66           & 67.13 ± 13.73          \\ 
                                    & \textbf{5}                        & 77.35 ± 14.41          & 76.36 ± 15.62          & \textbf{69.35 ± 11.79}  & \textbf{67.54 ± 13.86} \\ 
                                    & \textbf{7}                        & \textbf{77.64 ± 15.08} & \textbf{76.71 ± 16.22} & 68.89 ± 11.86           & 67.09 ± 13.93          \\ 
                                    & 9                                 & 76.85 ± 14.74          & 75.67 ± 16.46          & 68.72 ± 11.91           & 66.71 ± 14.17          \\ \midrule[0.1em]
\multirow{5}{*}{BNCI2015-001}      & \textbf{2}                        & \textbf{80.03 ± 13.17} & \textbf{79.25 ± 14.30} & 65.32 ± 11.18           & 61.60 ± 14.23          \\ 
                                    & 3                                 & 78.73 ± 13.73          & 77.68 ± 15.30          & 65.12 ± 11.02           & 61.21 ± 14.64          \\  
                                    & \textbf{5}                        & 79.03 ± 13.29          & 77.97 ± 14.93          & \textbf{66.22 ± 11.70}  & \textbf{62.55 ± 14.95} \\ 
                                    & 7                                 & 78.34 ± 13.66          & 76.72 ± 16.17          & 65.77 ± 10.94           & 62.16 ± 13.78          \\ 
                                    & 9                                 & 77.47 ± 13.68          & 75.78 ± 16.38          & 64.93 ± 11.36           & 61.05 ± 14.60          \\ \midrule[0.1em]
\multirow{5}{*}{SMR-BCI}           & \textbf{2}                        & \textbf{76.26 ± 16.53} & \textbf{75.17 ± 17.99} & 64.86 ± 13.74           & 60.03 ± 17.86          \\  
                                    & 3                                 & 75.83 ± 16.47          & 74.36 ± 18.42          & 66.14 ± 13.95           & 62.06 ± 17.69          \\  
                                    & \textbf{5}                        & 75.40 ± 16.33          & 74.14 ± 17.95          & \textbf{66.14 ± 14.00}  & \textbf{62.14 ± 17.73} \\  
                                    & 7                                 & 75.88 ± 16.34          & 74.65 ± 17.87          & 65.33 ± 14.39           & 61.12 ± 18.36          \\  
                                    & 9                                 & 75.57 ± 16.49          & 74.47 ± 18.01          & 64.60 ± 14.65           & 59.61 ± 19.26          \\ \midrule[0.1em]
\multirow{5}{*}{HighGamma}          & 2                                 & 78.98 ± 16.19          & 78.43 ± 16.78          & 68.25 ± 11.68           & 65.67 ± 14.34          \\ 
                                    & 3                                 & 79.70 ± 16.44          & 79.12 ± 17.10          & 69.54 ± 11.04           & 67.54 ± 13.31          \\ 
                                    & \textbf{5}                        & \textbf{80.23 ± 15.28} & \textbf{79.87 ± 15.67} & \textbf{69.84 ± 10.82}  & \textbf{67.97 ± 12.98} \\ 
                                    & 7                                 & 79.95 ± 15.47          & 79.52 ± 15.86          & 69.45 ± 11.23           & 67.56 ± 13.30          \\  
                                    & 9                                 & 79.55 ± 15.55          & 79.04 ± 16.08          & 68.48 ± 10.76           & 66.43 ± 13.12          \\ \midrule[0.1em]
\multirow{5}{*}{OpenBMI}            & 2                                 & 68.57 ± 16.79          & 67.74 ± 17.50          & 71.83 ± 14.29           & 70.93 ± 15.16          \\ 
                                    & 3                                 & 68.52 ± 17.12          & 67.59 ± 17.89          & 71.69 ± 14.33           & 70.87 ± 15.03          \\ 
                                    & \textbf{5}                        & \textbf{68.95 ± 16.75} & \textbf{68.04 ± 17.54} & \textbf{71.95 ± 14.18}  & \textbf{71.13 ± 14.88} \\ 
                                    & 7                                 & 69.09 ± 16.86          & 68.41 ± 17.37          & 71.52 ± 14.33           & 70.78 ± 14.91          \\  
                                    & 9                                 & 69.16 ± 16.45          & 68.37 ± 17.13          & 71.59 ± 14.20           & 70.81 ± 14.83          \\ \bottomrule[0.2em]
\end{tabular}}
\end{table}

\begin{table*}[t]
 \caption{Classification performance (Accuracy $\pm$ SD, F1-score $\pm$ SD, and AUC $\pm$ SD) in \% for the subject-dependent and subject-independent MI classification (using high-density EEG) on BCIC IV 2a, BNCI2015-001, SMR-BCI, High-Gamma, and OpenBMI compared to six different methods. The numerical values where MixNet's performance surpasses that of all the baseline approaches are denoted in bold.}
    \label{tab:result_binary_classification_20chs}
    \centering
    \small
    \resizebox{1.5\columnwidth}{!}{%
\begin{tabular}{@{}cccccccc@{}}
\toprule[0.2em]
\multirow{2}{*}{\textbf{Dataset}} & \multirow{2}{*}{\textbf{Models}} & \multicolumn{3}{c}{\textbf{Subject-dependent}}                           & \multicolumn{3}{c}{\textbf{Subject-independent}}                         \\ \cmidrule[0.1em](l){3-5} \cmidrule[0.1em](l){6-8}
                                  &                                  & \textbf{Accuracy}      & \textbf{F1-score}      & \textbf{AUC}           & \textbf{Accuracy}      & \textbf{F1-score}      & \textbf{AUC}           \\ \midrule[0.1em]
\multirow{6}{*}{BCIC IV 2a}           & FBCSP-SVM                        & 75.93 ± 14.93          & 74.60 ± 16.98          & 80.29 ± 23.75          & 58.06 ± 9.85           & 52.97 ± 13.87          & 62.21 ± 16.17          \\  
                                  & EEGNet-8,2                           & 65.93 ± 18.44          & 61.81 ± 22.41          & 69.83 ± 21.33          & 64.26 ± 11.03          & 61.32 ± 13.28          & 73.35 ± 14.07          \\  
                                  & DeepConvNet                      & 63.72 ± 17.18          & 62.66 ± 18.09          & 67.21 ± 20.40          & 56.34 ± 8.86           & 47.25 ± 14.27          & 66.50 ± 17.48          \\ 
                                  & Spectral-spatial CNN               & 76.91 ± 13.75          & 75.93 ± 14.63          & 86.77 ± 13.70          & 66.05 ± 13.70          & 63.18 ± 15.97          & 75.27 ± 16.53          \\ 
                                  & MIN2Net                          & 65.23 ± 16.14          & 64.68 ± 16.58          & 70.45 ± 19.97          & 60.03 ± 9.24           & 55.75 ± 12.84          & 68.61 ± 12.22          \\ 
                                  & MixNet                           & \textbf{77.64 ± 15.08}          & \textbf{76.71 ± 16.22} & 85.44 ± 15.02          & \textbf{69.35 ± 11.79} & \textbf{67.54 ± 13.86} & \textbf{77.49 ± 12.06} \\ \midrule[0.1em]
\multirow{6}{*}{BNCI2015-001}     & FBCSP-SVM                        & 78.36 ± 14.10          & 77.40 ± 15.21          & 76.18 ± 28.68          & 57.13 ± 10.58          & 48.70 ± 15.30          & 53.31 ± 28.76          \\ 
                                  & EEGNet-8,2                           & 67.41 ± 18.03          & 65.51 ± 19.87          & 71.76 ± 20.36          & 61.78 ± 12.16          & 56.59 ± 16.26          & 71.35 ± 16.90          \\ 
                                  & DeepConvNet                      & 70.55 ± 18.18          & 69.16 ± 19.60          & 75.56 ± 21.17          & 61.16 ± 13.94          & 54.82 ± 18.26          & 72.37 ± 18.65          \\ 
                                  & Spectral-spatial CNN               & 79.82 ± 12.55          & 78.61 ± 13.95          & 90.82 ± 9.21           & 63.77 ± 13.74          & 57.10 ± 18.84          & 78.58 ± 18.29          \\ 
                                  & MIN2Net                          & 74.06 ± 16.13          & 73.45 ± 16.74          & 79.27 ± 18.25          & 60.75 ± 8.17           & 56.98 ± 10.90          & 70.26 ± 13.58          \\  
                                  & MixNet                           & \textbf{80.03 ± 13.17} & \textbf{79.25 ± 14.30} & 87.88 ± 11.50          & \textbf{66.22 ± 11.70} & \textbf{62.55 ± 14.95} & 78.56 ± 15.76          \\ \midrule[0.1em]
\multirow{6}{*}{SMR-BCI}          & FBCSP-SVM                        & 74.19 ± 17.28          & 72.70 ± 18.79          & 67.39 ± 21.61          & 62.71 ± 15.42          & 55.47 ± 21.23          & 65.59 ± 24.24          \\  
                                  & EEGNet-8,2                           & 67.76 ± 18.09          & 64.55 ± 21.33          & 71.79 ± 21.40          & 58.07 ± 11.45          & 48.24 ± 17.04          & 72.52 ± 18.35          \\  
                                  & DeepConvNet                      & 61.40 ± 15.66          & 59.28 ± 17.17          & 63.58 ± 19.12          & 65.26 ± 16.83          & 59.03 ± 22.02          & 76.48 ± 20.18          \\  
                                  & Spectral-spatial CNN               & 76.76 ± 16.66          & 74.39 ± 20.39          & 83.07 ± 16.86          & 66.21 ± 15.15          & 60.21 ± 20.61          & 78.94 ± 17.73          \\  
                                  & MIN2Net                          & 65.90 ± 16.50          & 64.98 ± 16.96          & 70.87 ± 19.83          & 59.79 ± 13.72          & 51.17 ± 19.07          & 77.53 ± 19.60          \\ 
                                  & MixNet                           & 76.26 ± 16.53          & \textbf{75.17 ± 17.99} & 82.37 ± 17.65          & 66.14 ± 14.00          & \textbf{62.14 ± 17.73} & 75.76 ± 17.27          \\ \midrule[0.1em]
\multirow{6}{*}{HighGamma}        & FBCSP-SVM                        & 76.29 ± 16.35          & 75.98 ± 16.66          & 76.45 ± 19.77          & 62.86 ± 10.87          & 58.65 ± 14.17          & 72.21 ± 14.73          \\  
                                  & EEGNet-8,2                           & 70.64 ± 19.20          & 69.67 ± 20.23          & 74.87 ± 20.04          & 63.52 ± 11.44          & 59.59 ± 14.64          & 74.91 ± 14.93          \\ 
                                  & DeepConvNet                      & 71.05 ± 19.95          & 69.04 ± 22.02          & 75.49 ± 22.87          & 66.82 ± 12.49          & 63.19 ± 16.09          & 78.72 ± 15.15          \\  
                                  & Spectral-spatial CNN               & 77.89 ± 14.69          & 76.91 ± 16.00          & 84.25 ± 13.56          & 65.29 ± 12.49          & 61.01 ± 16.31          & 76.91 ± 13.60          \\  
                                  & MIN2Net                          & 73.48 ± 18.23          & 73.33 ± 18.35          & 77.77 ± 19.56          & 68.60 ± 11.79          & 66.83 ± 13.26          & 77.74 ± 15.89          \\ 
                                  & MixNet                           & \textbf{80.23 ± 15.28} & \textbf{79.87 ± 15.67} & \textbf{85.90 ± 15.15} & \textbf{70.00 ± 9.53}  & \textbf{68.71 ± 10.58} & \textbf{79.60 ± 12.78} \\ \midrule[0.1em]
\multirow{6}{*}{OpenBMI}          & FBCSP-SVM                        & 66.20 ± 16.43          & 65.12 ± 17.32          & 59.75 ± 25.22          & 64.96 ± 12.70          & 62.99 ± 14.24          & 70.82 ± 16.64          \\  
                                  & EEGNet-8,2                           & 60.41 ± 17.12          & 57.77 ± 19.04          & 62.19 ± 19.58          & 68.84 ± 13.87          & 66.48 ± 16.36          & 76.77 ± 15.06          \\  
                                  & DeepConvNet                      & 60.31 ± 16.76          & 58.60 ± 17.85          & 62.10 ± 19.03          & 68.33 ± 15.33          & 65.71 ± 17.80          & 75.55 ± 16.96          \\  
                                  & Spectral-spatial CNN               & 65.19 ± 15.94          & 62.20 ± 19.19          & 69.36 ± 17.47          & 68.24 ± 13.54          & 67.06 ± 14.45          & 75.06 ± 16.17          \\ 
                                  & MIN2Net                          & 61.03 ± 14.47          & 59.78 ± 15.29          & 64.56 ± 17.52          & 72.03 ± 14.04          & 71.17 ± 14.80          & 78.92 ± 14.97          \\ 
                                  & MixNet                           & \textbf{68.95 ± 16.75} & \textbf{68.04 ± 17.54} & \textbf{73.30 ± 18.44} & 71.95 ± 14.18          & 71.13 ± 14.88          & 78.05 ± 15.97          \\ \bottomrule[0.2em]
\end{tabular}}
\end{table*}

\begin{table*}[t]
 \caption{Classification performance (Accuracy $\pm$ SD, F1-score $\pm$ SD, and AUC $\pm$ SD) in \% for the subject-dependent and subject-independent MI classification (using 3-channel EEG) on BCIC IV 2b dataset compared to six different methods. Bold denotes the best MixNet's performance when it outperforms all the baseline methods.}
    \label{tab:result_binary_classification_3chs}
    \centering
    \resizebox{1.5\columnwidth}{!}{%
\begin{tabular}{@{}cccccccc@{}}
\toprule[0.2em]
\multirow{2}{*}{\textbf{Dataset}} & \multirow{2}{*}{\textbf{Models}} & \multicolumn{3}{c}{\textbf{Subject-dependent}}                  & \multicolumn{3}{c}{\textbf{Subject-independent}}                         \\  \cmidrule[0.1em](l){3-5} \cmidrule[0.1em](l){6-8}
                                  &                                  & \textbf{Accuracy}      & \textbf{F1-score}      & \textbf{AUC}  & \textbf{Accuracy}      & \textbf{F1-score}      & \textbf{AUC}           \\ \midrule[0.1em]
\multirow{6}{*}{BCIC2b}           & FBCSP-SVM                        & 74.80 ± 14.27          & 74.30 ± 14.63          & 78.91 ± 20.23 & 66.31 ± 10.15          & 63.68 ± 12.07          & 73.70 ± 16.52          \\  
                                  & EEGNet-8,2                           & 73.91 ± 14.00          & 72.10 ± 16.63          & 81.70 ± 15.85 & 72.46 ± 10.34          & 71.48 ± 11.15          & 82.32 ± 14.10          \\ 
                                  
                                  & DeepConvNet                      & 75.30 ± 15.29          & 73.95 ± 16.64          & 84.06 ± 17.56 & 73.96 ± 12.11          & 72.91 ± 12.97          & 83.78 ± 14.26          \\ 
                                  
                                  & Spectral-spatial CNN               & 75.99 ± 13.51          & 75.16 ± 14.34          & 85.29 ± 14.77 & 72.72 ± 10.80          & 71.06 ± 13.25          & 83.16 ± 12.96          \\ 
                                  & MIN2Net                          & 74.55 ± 14.45          & 74.23 ± 14.56          & 81.52 ± 16.80 & 74.67 ± 10.90          & 74.17 ± 11.14          & 83.43 ± 14.30          \\ 
                                  & MixNet                           & \textbf{77.07 ± 14.59} & \textbf{76.84 ± 14.68} & 83.90 ± 16.92 & \textbf{75.66 ± 10.49} & \textbf{75.23 ± 10.78} & \textbf{84.03 ± 12.93} \\ \bottomrule[0.2em]
\end{tabular}}
\end{table*}

\section{Discussion}

MixNet is developed in this paper to address the unstable learning behavior encountered in MIN2Net, particularly evident when dealing with limited data sizes and the need for exhaustive parameter searches to determine the optimal set of loss weights for the multi-task learning architecture. The effectiveness of MixNet is demonstrated through its performances, as presented in both \autoref{tab:result_binary_classification_20chs} and \autoref{tab:result_binary_classification_3chs}. MixNet indicates outstanding performance across various data sizes, including small, medium, and large, in both subject-dependent and subject-independent settings. Notably, MixNet achieved the highest F1-score for all evaluated datasets in both settings, indicating its effectiveness in reducing false positives and negatives. Furthermore, MixNet consistently outperforms all state-of-the-art methods in terms of accuracy, F1-score, and AUC in both scenarios, particularly with large-scale datasets such as the High-Gamma and OpenBMI datasets. This finding exhibits the importance of the volume of training samples in enhancing MixNet's classification efficacy, demonstrating its robustness with massive training data.

Based on the findings from \autoref{tab:adaptive_min2net_vary_csp_components} and \autoref{tab:AE_triplet_margin_results}, it is obvious that the optimal number of spatial filters and margins vary significantly between subject-dependent and subject-independent settings. The difference can be explained by the substantial differences in the distribution of EEG data between the two settings. In other words, there is a high variability observed in EEG signals among different subjects when considering subject-independent analysis, whereas this issue does not arise when considering subject-dependent analysis. The margin values in the subject-independent are greater than those in the subject-dependent. This is because the margin must be present to ensure that the learned embeddings have a structured space where similar samples are clustered together and there is a clear differentiation between different classes.

In contrast, \autoref{tab:AE_triplet_latent_dims_results} demonstrates that configuring MixNet with latent vector dimensions $z = U \times N_{b}$ results in optimal performance in both subject-dependent and subject-independent scenarios across the SMR-BCI, High-Gamma, and OpenBMI datasets. This finding suggests that setting the latent vector sizes equal to MixNet's input dimensions enables MixNet to capture generalized features and provide optimal MI classification performance.

\begin{figure}[t]
\centering
\includegraphics[width=1.00\columnwidth]{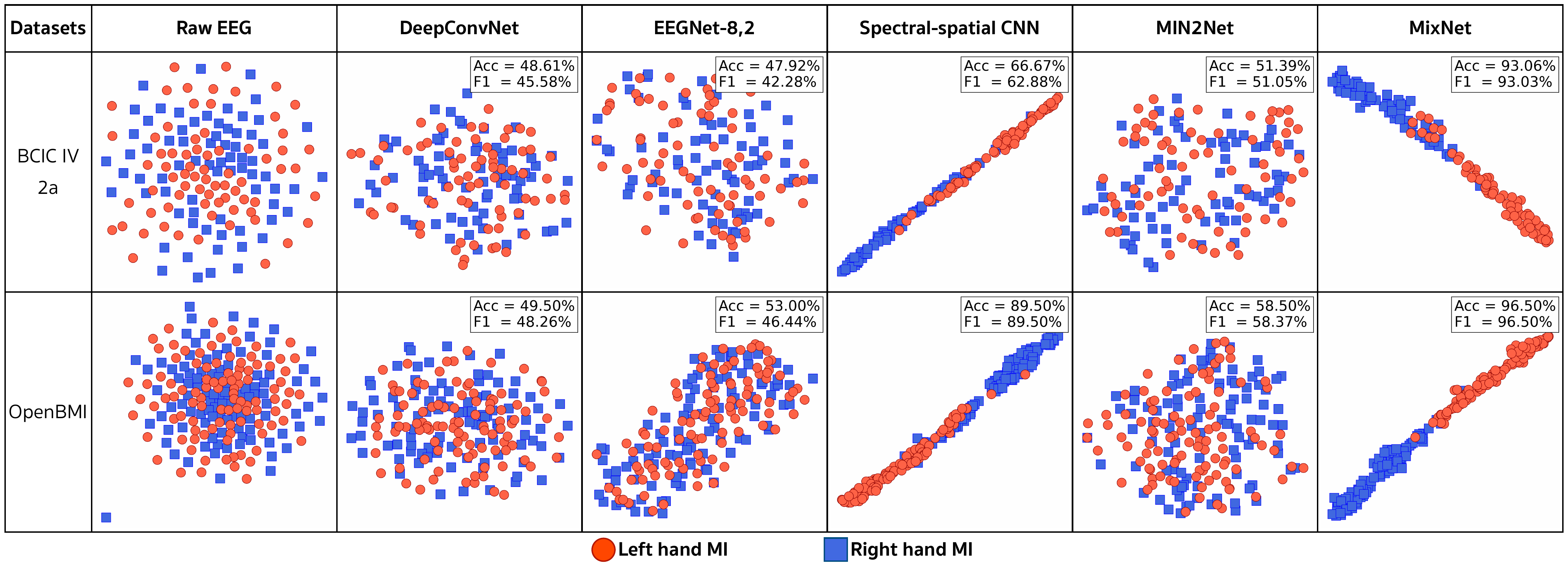}
\caption{Representation of raw and learned EEG features generated by each considered method using $t$-SNE projection. The image compares two-dimensional $t$-SNE projections designed for subject-dependent binary classification.}
\label{fig:sub_dependent_tsne_two_classes_mixnet_work}
\end{figure}

Based on the findings illustrated in \autoref{fig:sub_dependent_tsne_two_classes_mixnet_work}, it can be seen that the latent embedding features produced by MixNet exhibit a higher degree of compactness in the setting of subject-dependent MI classification, in comparison to the other baseline methods. Furthermore, they are extending their presence across the projection space for each class. The finding indicates that MixNet has superior results in the learning process when the data size is either large or small, owing to the high-quality representation of the MI signals in the learned latent embedding features.

\begin{figure}[t]
\centering
\includegraphics[width=1.00\columnwidth]{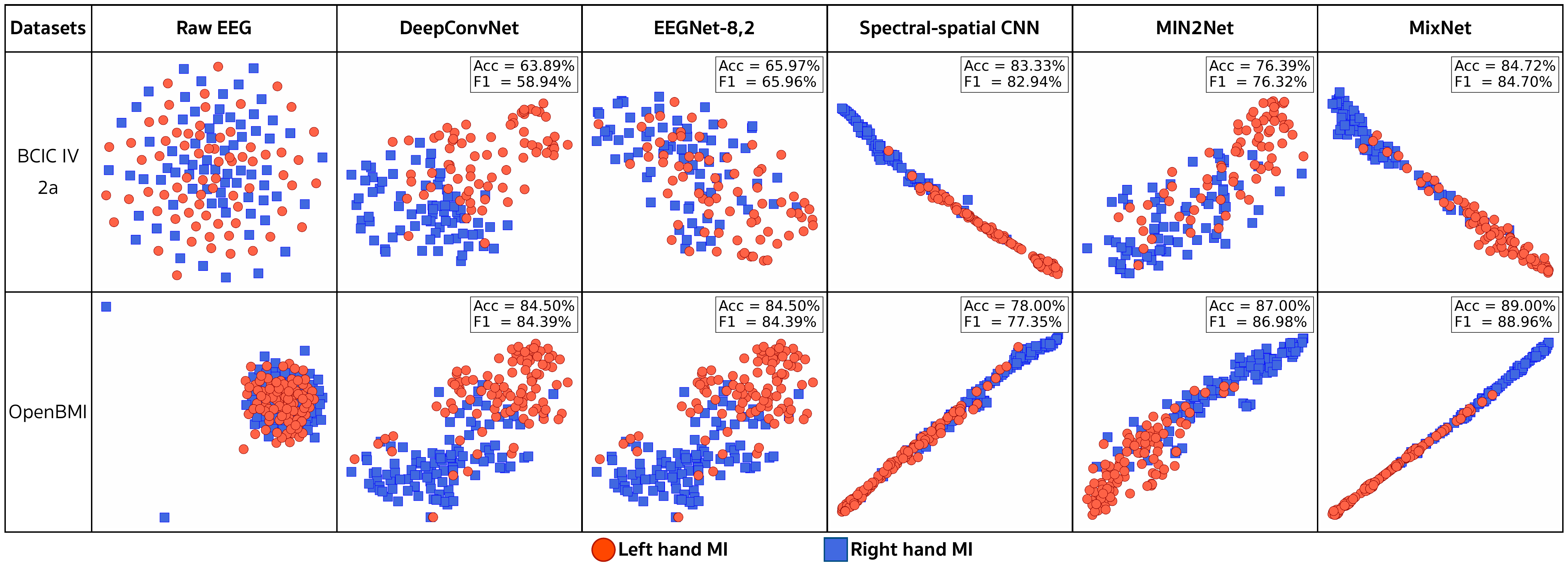}
\caption{Representation of raw and learned EEG features generated by each considered method using $t$-SNE projection. The picture compares two-dimensional $t$-SNE projections designed for subject-independent binary classification.}
\label{fig:sub_independent_tsne_two_classes_mixnet_work}
\end{figure}

In a subject-independent setting, as illustrated in \autoref{fig:sub_independent_tsne_two_classes_mixnet_work}, MixNet generates highly discriminative patterns over the small- and large-scale datasets (i.e., BCIC IV 2a and OpenBMI, respectively) compared to the other baseline methods. Consequently, MixNet outperforms others due to its superior MI representation in the learned latent embedding features.

\begin{figure}[t]
\centering
\includegraphics[width=1.00\columnwidth]{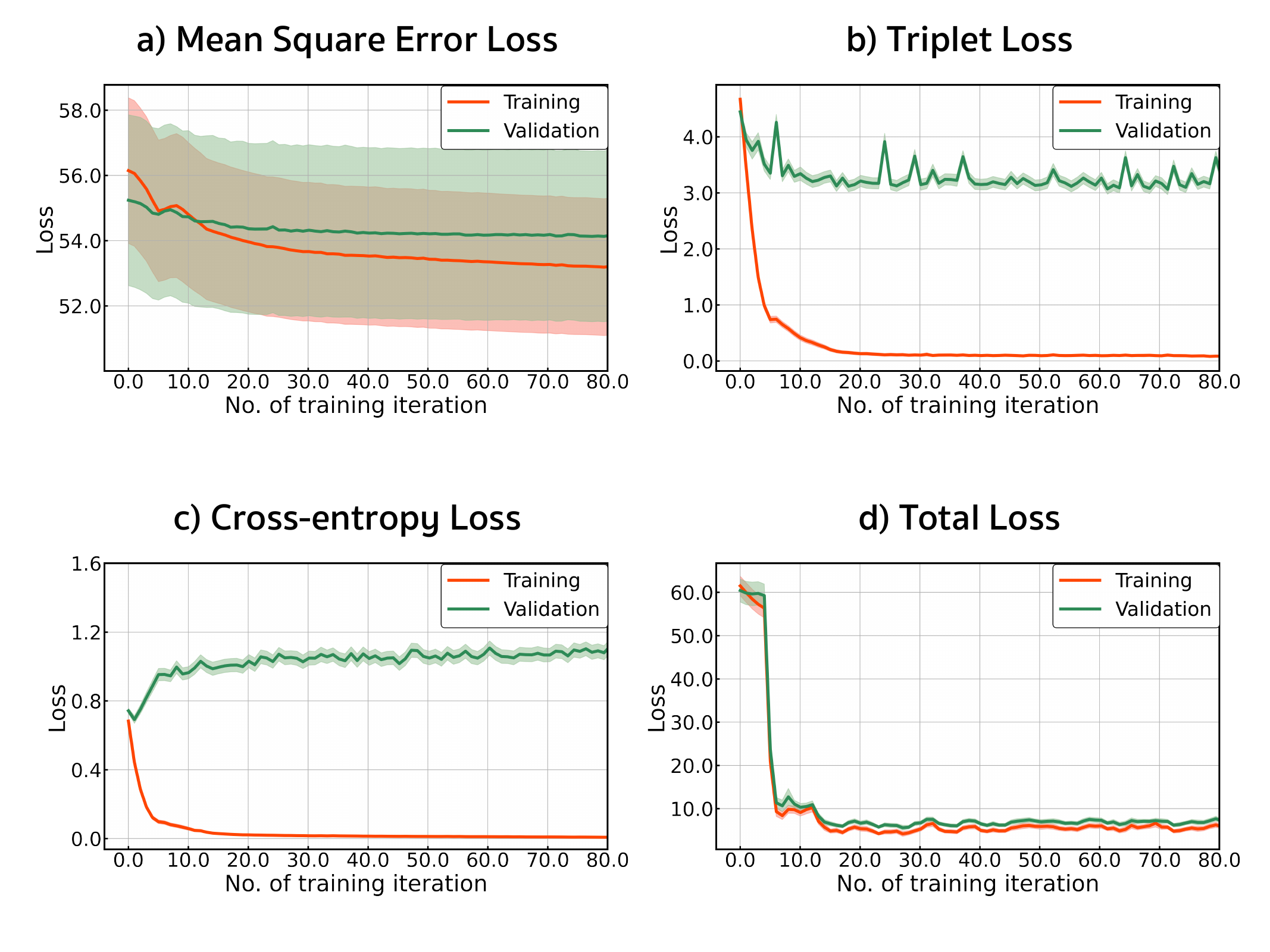}
\caption{The training and validation losses during the OpenBMI dataset's training period for subject-dependent binary classification. The graphs depict the averaged losses for 54 subjects, including a) mean square error, b) triplet loss, c) cross-entropy loss, and d) total loss.}
\label{fig:visualize_loss_subject_dependent}
\end{figure}

\begin{figure}[t]
\centering
\includegraphics[width=1.00\columnwidth]{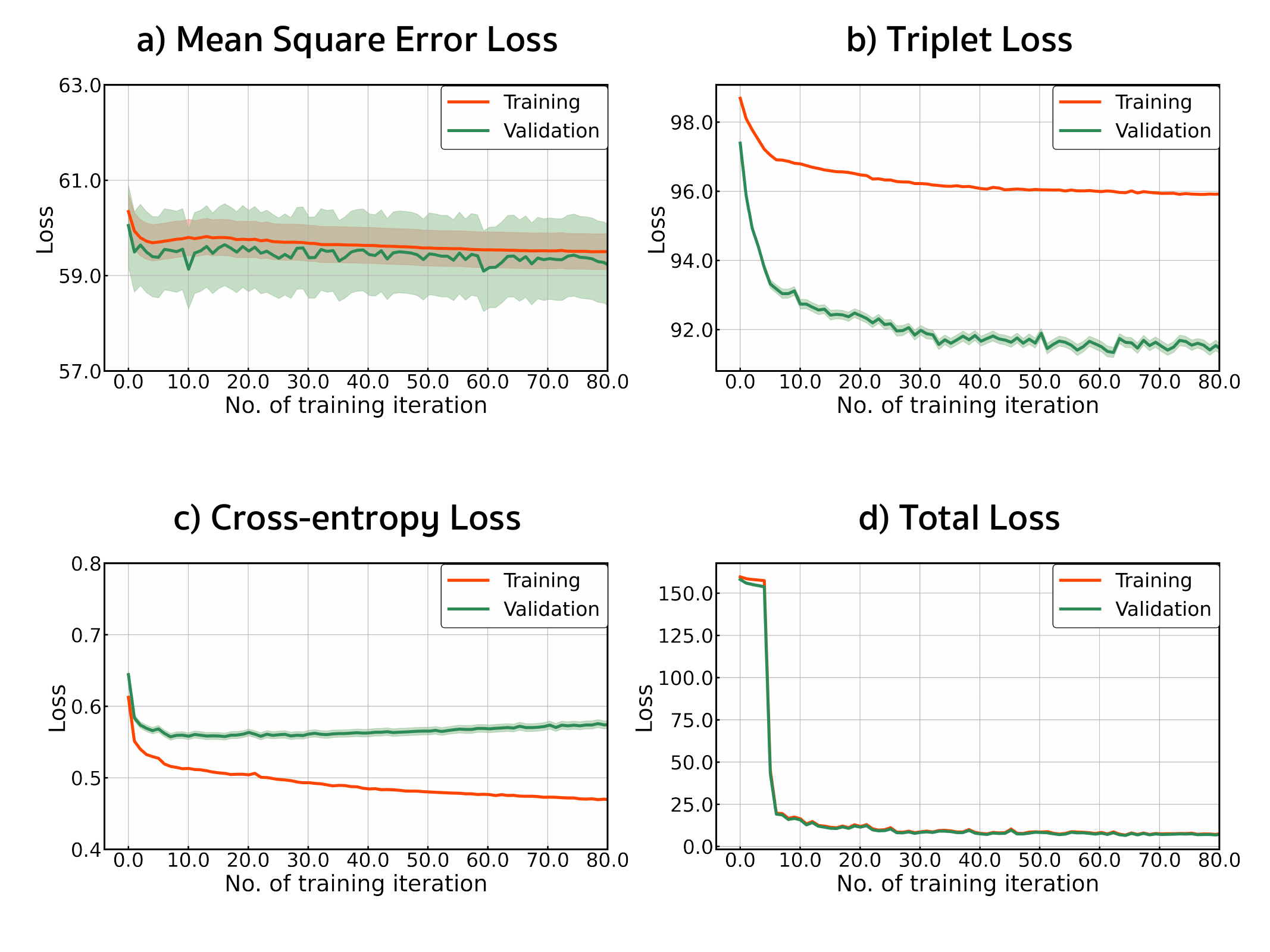}
\caption{The training and validation losses during the OpenBMI dataset's training period for subject-independent binary classification. The plots exhibit the averaged losses for 54 subjects, including a) mean square error, b) triplet loss, c) cross-entropy loss, and d) total loss.}
\label{fig:visualize_loss_subject_independent}
\end{figure}

To gain a deeper understanding of the internal workings of the optimization process of MixNet, we analyzed the changes in training and validation losses for binary classification. This analysis was performed on the OpenBMI dataset, utilizing both subject-dependent (\autoref{fig:visualize_loss_subject_dependent}) and subject-independent (\autoref{fig:visualize_loss_subject_independent}) manners. The study monitored four distinct types of losses in the MixNet model: mean squared error (MSE), triplet loss, cross-entropy (CE) loss, and total loss. This monitoring was conducted throughout 80 epochs for all subjects.

In the subject-dependent task, it can be observed that three validation losses did not exhibit signs of overfitting, except for the cross-entropy (CE) loss, which showed indications of rapid overfitting. This suggests that the supervised learning module in MixNet might be less effective when dealing with small data sizes. However, the unsupervised and deep metric learning modules play vital roles in helping MixNet alleviate overfitting during its training. As a result, MixNet's total loss continues to perform well without showing signs of overfitting. In the subject-independent task, it can be seen that the convergence process of MixNet in this learning task is remarkably robust. We noticed that the cross-entropy (CE) loss starts displaying signs of overfitting after around ten epochs, whereas the other losses do not show any indications of overfitting.

\begin{figure}[t]
  \centering
  \subfloat[\label{fig:loss_weights_subject_dependent}]{\includegraphics[width=0.5\columnwidth]{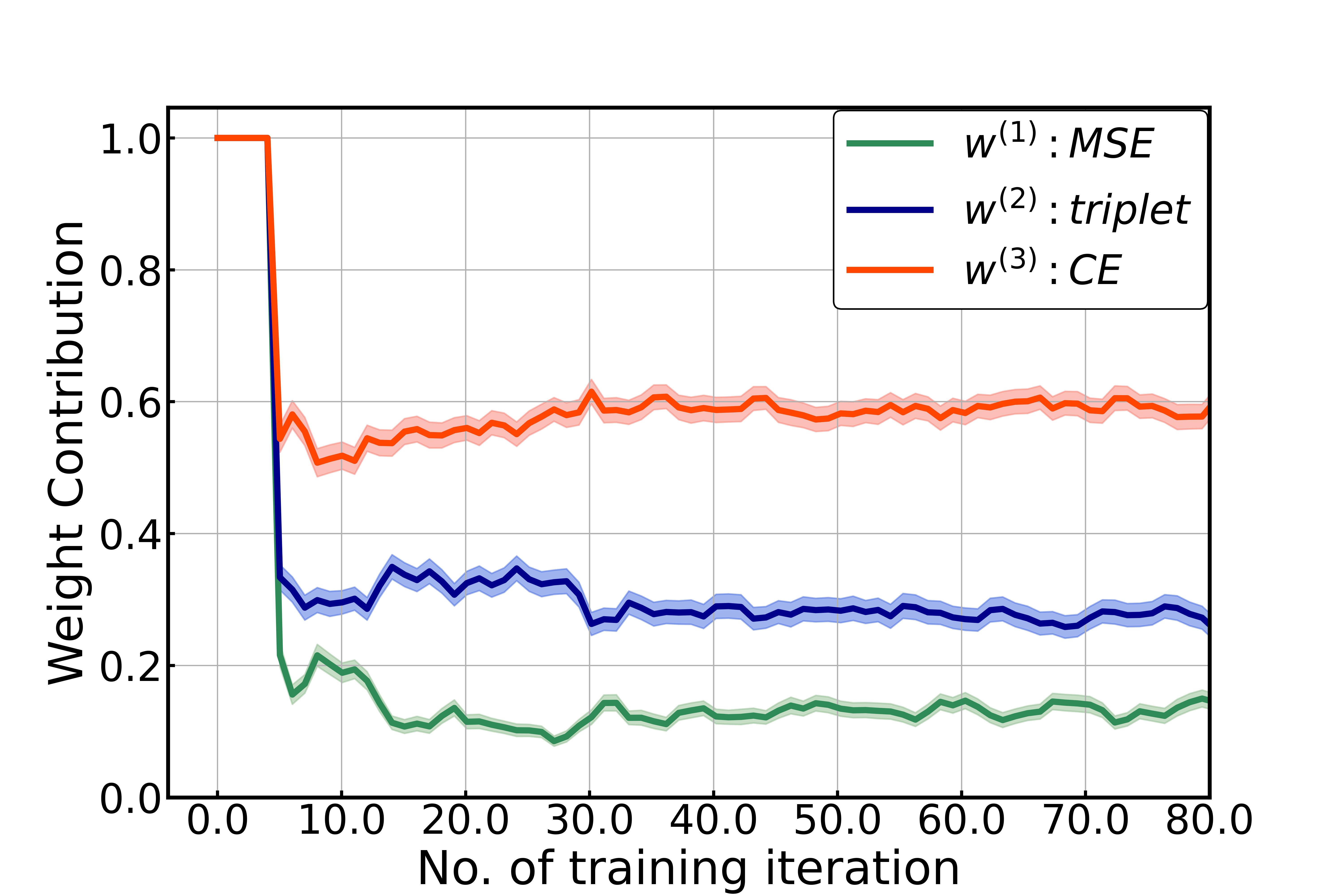}} 
  \subfloat[\label{fig:loss_weights_subject_independent}]{\includegraphics[width=0.5\columnwidth]{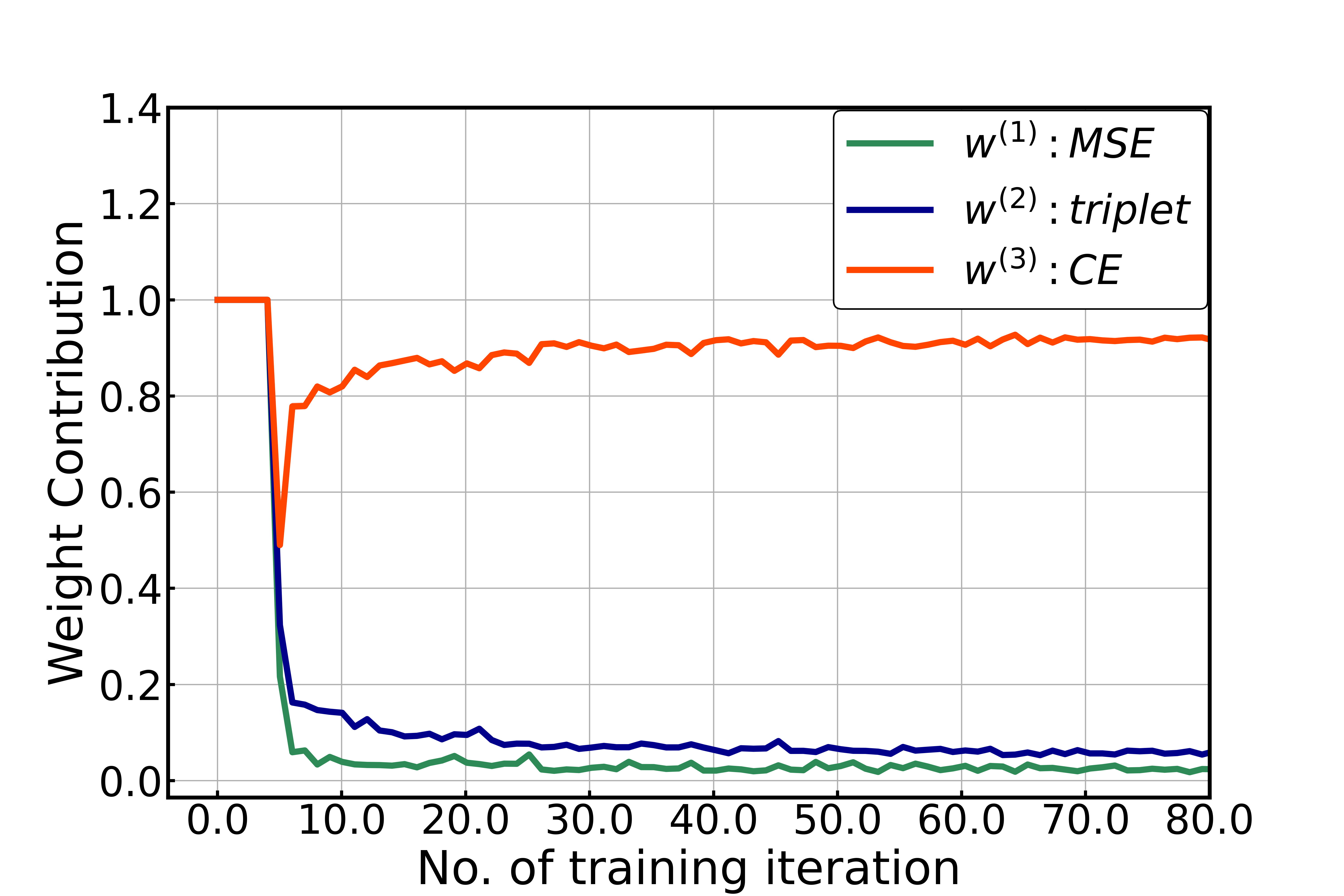}}
  \caption{The change of the adaptive loss weights during the OpenBMI dataset's training period, specifically for binary classification. The graphs illustrate the averaged loss weights for all cross-validation folds using 54 subjects, including (a) the subject-dependent scheme and (b) the subject-independent scheme.}
  \label{fig:visualize_loss_weights}
\end{figure}

To observe how MixNet's loss weights ($w^{(1)}$, $w^{(2)}$, and $w^{(3)}$) were adapted for adaptive gradient blending during MixNet's training, we examined the progression of adaptive loss weights throughout all cross-validation folds on the OpenBMI dataset, as depicted in \autoref{fig:visualize_loss_weights}. The graphs exhibit that all weights decrease rapidly, which helps to impede learning and prevent regular overfitting of the data. Interestingly, the weight ratio for the cross-entropy (CE) loss in subject-dependent learning is much lower than in subject-independent learning. This indicates that reducing the cross-entropy loss weight in a multi-tasking architecture while increasing other loss weights, as seen in MixNet's subject-dependent classification, can help alleviate overfitting when dealing with a small size of data.

To further assess the feasibility of MixNet for online BCI systems, we conducted an experiment using the OpenBMI and BCIC IV 2b datasets, comparing MixNet's time complexity with that of established baseline methods. According to the results shown in \autoref{tab:result_time}, MixNet has lower trainable parameters than Spectral-spatial CNN and DeepConvNet for the low-density EEG-MI dataset and lower than Spectral-spatial CNN for the high-density EEG-MI dataset. MixNet's GPU memory usage is significantly lower than that of DeepConvNet and Spectral-spatial CNN for both datasets, although it is slightly higher than that of MIN2Net and EEGNet. Additionally, MixNet demonstrates training and prediction speeds slightly different from those of compact baseline models, such as EEGNet-8,2 and DeepConvNet. These findings suggest that MixNet could be effectively used in online BCI systems because it offers compactness and maintains latency or prediction times below 0.5 seconds during testing sessions.

While MixNet demonstrates promising classification results, there remains plenty of room for further enhancement. EEG multi-class classification poses a significant challenge for MixNet due to the requirement of transforming the inputs using FBCSP. However, this transformation technique does not sufficiently capture the characteristics of multi-class EEG data. Finally, we can investigate the application of MixNet to additional EEG measurements, including steady-state visual evoked potentials (SSVEP), movement-related cortical potentials (MRCPs), and event-related potentials (ERP). MixNet can help obtain the most distinguishing features for classification.

\begin{table}[t]
    \caption{Time complexity of training (\textit{T\textsubscript{train}}) and prediction (\textit{T\textsubscript{pred}}) in seconds per epoch for all methods,  number of trainable parameters, and GPU memory usage in (MB) for all approaches on low- and high-density EEG-MI datasets.}
    
    \label{tab:result_time}
    \centering
    \small
    \resizebox{1.0\columnwidth}{!}{%
        \begin{tabular}{@{}cccccccc@{}}
        \toprule[0.2em]
        \multirow{2}{*}{\textbf{Dataset}} & \multirow{2}{*}{\textbf{Comparison Model}} & \multirow{2}{*}{\textit{\textbf{\begin{tabular}[c]{@{}c@{}}\# trainable\\ params\end{tabular}}}} & \multirow{2}{*}{\textit{\textbf{\begin{tabular}[c]{@{}c@{}}GPU Memory\\ Usage (MB)\end{tabular}}}} & \multicolumn{2}{c}{\textbf{Subject-dependent}}                  & \multicolumn{2}{c}{\textbf{Subject-independent}}                \\ \cmidrule[0.1em](l){5-6} \cmidrule[0.1em](l){7-8}  
                                        &                                      
                                        &                                        & 
                                        & \textbf{\textit{T\textsubscript{train}}} & \textbf{\textit{T\textsubscript{pred}}} & \textbf{\textit{T\textsubscript{train}}} & \textbf{\textit{T\textsubscript{pred}}} \\ \midrule[0.1em]
    \multirow{5}{*}{\begin{tabular}[c]{@{}c@{}}BCIC \\ IV 2b\end{tabular}}        & FBCSP-SVM                              & -                      & -                                       & -                                & 0.0043                       & -                                & 0.0405                       \\
                                        & EEGNet-8,2                            & 4,890         & 0.1218                                   & 0.1415                           & 0.2671                       & 0.6584                           & 0.2562                             \\                
                                        & DeepConvNet                          & 142,802        & 2.6230                               & 0.1568                           & 0.3120                       & 0.6223	                           & 0.3007                       \\

                                        & Spectral-Spatial CNN                  & 16,475,634    & 253.66015                             & 0.8178                         & 0.8753                       & 2.4189                           & 0.8377                       \\
                                        & MIN2Net                               & 10,709     & 0.8391                                & 0.2148	                           & 0.2875                       & 0.7857                           & 0.3052                       \\ 
                                        & MixNet                               & 44,777     & 1.3862                                 & 0.2573	                           & 0.3150                       & 1.2081                           & 0.3262 \\ \midrule[0.1em]

     \multirow{5}{*}{OpenBMI}           & FBCSP-SVM                             & -                                          & -                                & 0.0020                       & -                                & 0.1906                       \\ 
                                        & EEGNet-8,2                            & 5,162    & 0.1199                                    & 0.1882                           & 0.1439                       & 3.0951                           & 0.1372                       \\
                                        & DeepConvNet                          & 153,427   & 2.6079                                 & 0.1804                           & 0.1618                       & 1.7497                           & 0.4734                       \\

                                        & Spectral-Spatial CNN                  & 77,577,714   & 1210.0701                              & 2.2476                           & 1.0934                       & 11.9067                          & 0.8560                       \\
                                        & MIN2Net                               & 55,232     & 1.3086                                & 0.3527                           & 0.2851                       & 1.3626                           & 0.1043     
                                         \\
                                        & MixNet                               & 178,328    & 1.0491                                   & 0.1614	                           & 0.3160                       & 3.9670                           & 0.3201     
                                        \\ \bottomrule[0.2em]
        \end{tabular}
    }
\end{table}
\section{Conclusion}
This paper presents a novel framework known as MixNet, which builds upon the core ideas of MIN2Net and efficiently addresses difficulties faced in subject-specific learning, multi-task learning, and loss weight optimization. The development of MixNet involves the integration of spectral-spatial signals, adaptive gradient blending, and a multi-task autoencoder. The integration of this fusion allows MixNet to effectively compress and extract distinctive patterns from EEG-MI data while concurrently performing the classification of MI classes. To evaluate MixNet's performance, we conducted binary classification comparisons with five baseline methods across six benchmark datasets. The results of the classification analysis demonstrated that MixNet outperformed all state-of-the-art methods in subject-dependent and subject-independent MI classification tasks over six benchmark datasets in terms of the F1-score metric. Notably, the MI classification performance on the low-density EEG database BCIC IV 2b stands out among the six experimental datasets. The proposed method demonstrated promising experimental results in this context, outperforming state-of-the-art algorithms in classifying EEG-MI signals with three EEG channels. This finding indicates the possibility and practicability of using this model for developing a low-density EEG system, such as low-density wearable devices. This system could make deploying EEG systems in various real-world scenarios, such as home monitoring, remote healthcare, and mobile applications, more feasible and cost-effective.

\bibliographystyle{IEEEtran}
\bibliography{References}
\begin{IEEEbiography}[{\includegraphics[width=1in,height=1.25in, clip, keepaspectratio]{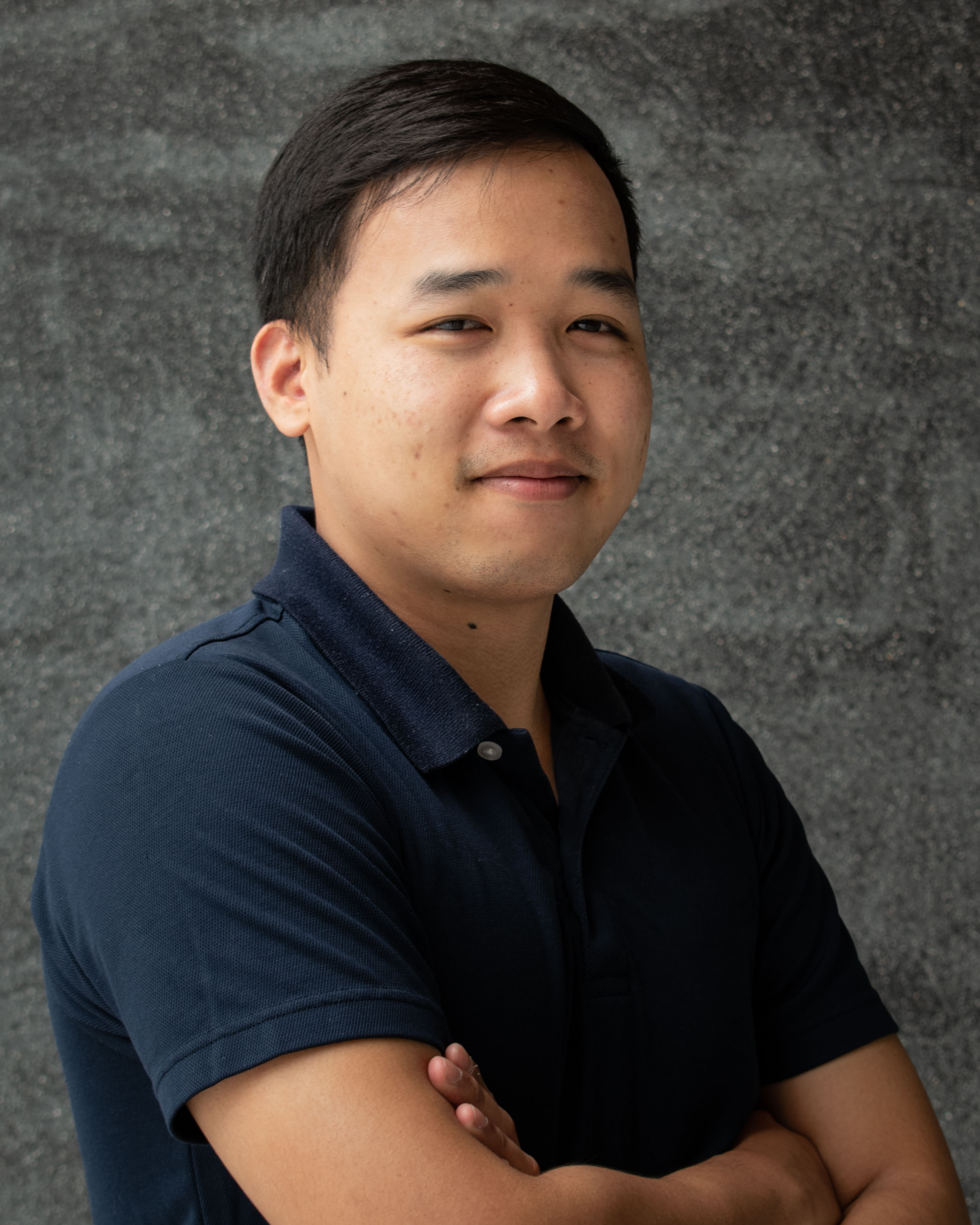}}]{Phairot Autthasan} received the B.Sc. degree in Chemistry Department, King Mongkut's University of Technology Thonburi, Bangkok, Thailand in 2017, and the Ph.D. degree in Information Science and Technology, Vidyasirimedhi Institute of Science and Technology (VISTEC), Thailand, in 2023. He was a visiting Ph.D. researcher in Department of Electrical Engineering (ESAT), Katholieke Universiteit Leuven (KU Leuven), Belgium, in 2022. He is currently a postdoctoral researcher at the Vidyasirimedhi Institute of Science and Technology, Thailand. His research interests include AI for healthcare, biosignal analysis, brain-computer interfaces, and machine learning for biosignals.
\end{IEEEbiography}
\vspace{-8mm}
\begin{IEEEbiography}[{\includegraphics[width=1in,height=1.25in,clip,keepaspectratio]{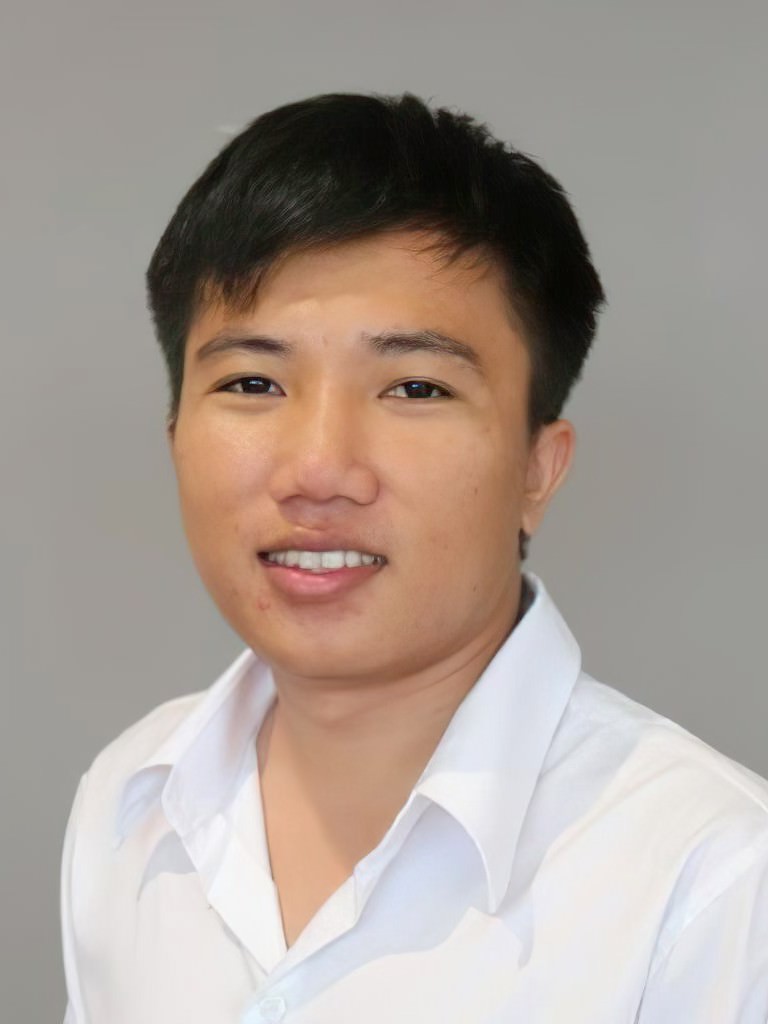}}]{Rattanaphon Chaisaen} received the B.Sc. degree in computer science from Khon Kaen University, Khon Kaen, Thailand, in 2018. He is currently working toward the Ph.D. degree in information science and technology with the Bio-Inspired Robotics and Neural Engineering Laboratory, School of Information Science and Technology, Vidyasirimedhi Institute of Science and Technology (VISTEC), Thailand. His research interests include deep learning approaches applied to biosignals, brain-computer interfaces, and assistive technology.
\end{IEEEbiography}
\vspace{-8mm}
\begin{IEEEbiography}[{\includegraphics[width=1in,height=1.25in,clip,keepaspectratio]{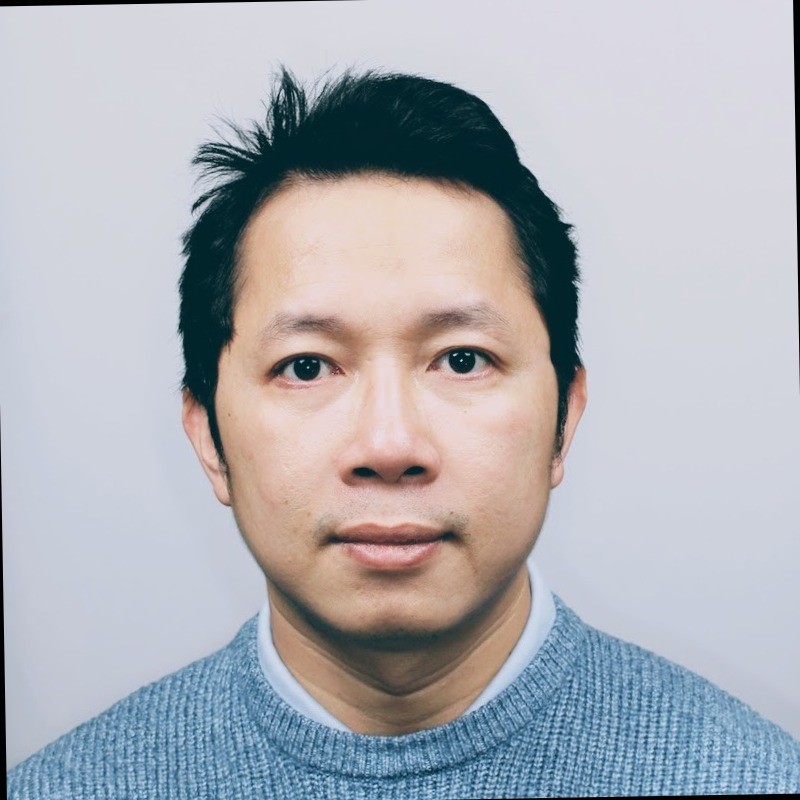}}]{Huy Phan} received the Dr.-Ing. degree (summa cum laude) in computer science from the University of Lübeck, Germany, in 2017. From 2017 to 2018, he was a postdoctoral research assistant with the University of Oxford, U.K. From 2019 to 2020, he was a lecturer with the University of Kent, U.K. From 2020-2022, he was a lecturer in artificial intelligence Queen Mary University of London, U.K. and a Turing Fellow at The Alan Turing Institute, U.K. In 2023, he joined Amazon where he is currently a senior research scientist. His research interests include machine learning and signal processing, with a focus on audio and biosignal analysis. He was the recipient of the Bernd Fischer Award for the best PhD thesis from the University of Lübeck in 2018 and the Benelux’s IEEE-EMBS Best Paper Award 2019-20. He is an IEEE senior member.
\end{IEEEbiography}
\vspace{-8mm}
\begin{IEEEbiography}[{\includegraphics[width=1in,height=1.25in,clip,keepaspectratio]{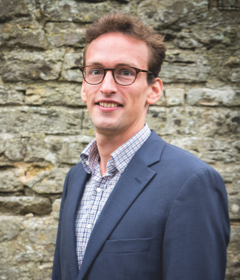}}]{Maarten De Vos} has a joint appointment as Professor in the Departments of Engineering and Medicine at KU Leuven after being Associate Professor at the University of Oxford, United Kingdom, and Junior Professor at the University of Oldenburg, Germany. He obtained an MSc (2005) and PhD (2009) in Electrical Engineering from KU Leuven, Belgium. His academic work focuses on AI for health, innovative biomedical monitoring and signal analysis for daily life applications, in particular the derivation of personalised biosignatures of patient health from data acquired via wearable sensors and the incorporation of smart analytics into unobtrusive systems. His pioneering research in the field of mobile real-life brain-monitoring has won several innovation prices, among which the prestigious Mobile Brain Body monitoring prize in 2017. In 2019, he was awarded the Martin Black Prize for the best paper in Physiological Measurements. In 2023, he was also elected as Laureate of the Flemish Academy of Sciences, discipline Technical Sciences. He is Associate Editor of Journal of Biomedical and Health Informatics, and on the editorial board of Journal of Neural Engineering and Nature Digital Medicine. 
\end{IEEEbiography}
\vspace{-8mm}
\begin{IEEEbiography}[{\includegraphics[width=1in,height=1.25in,clip,keepaspectratio]{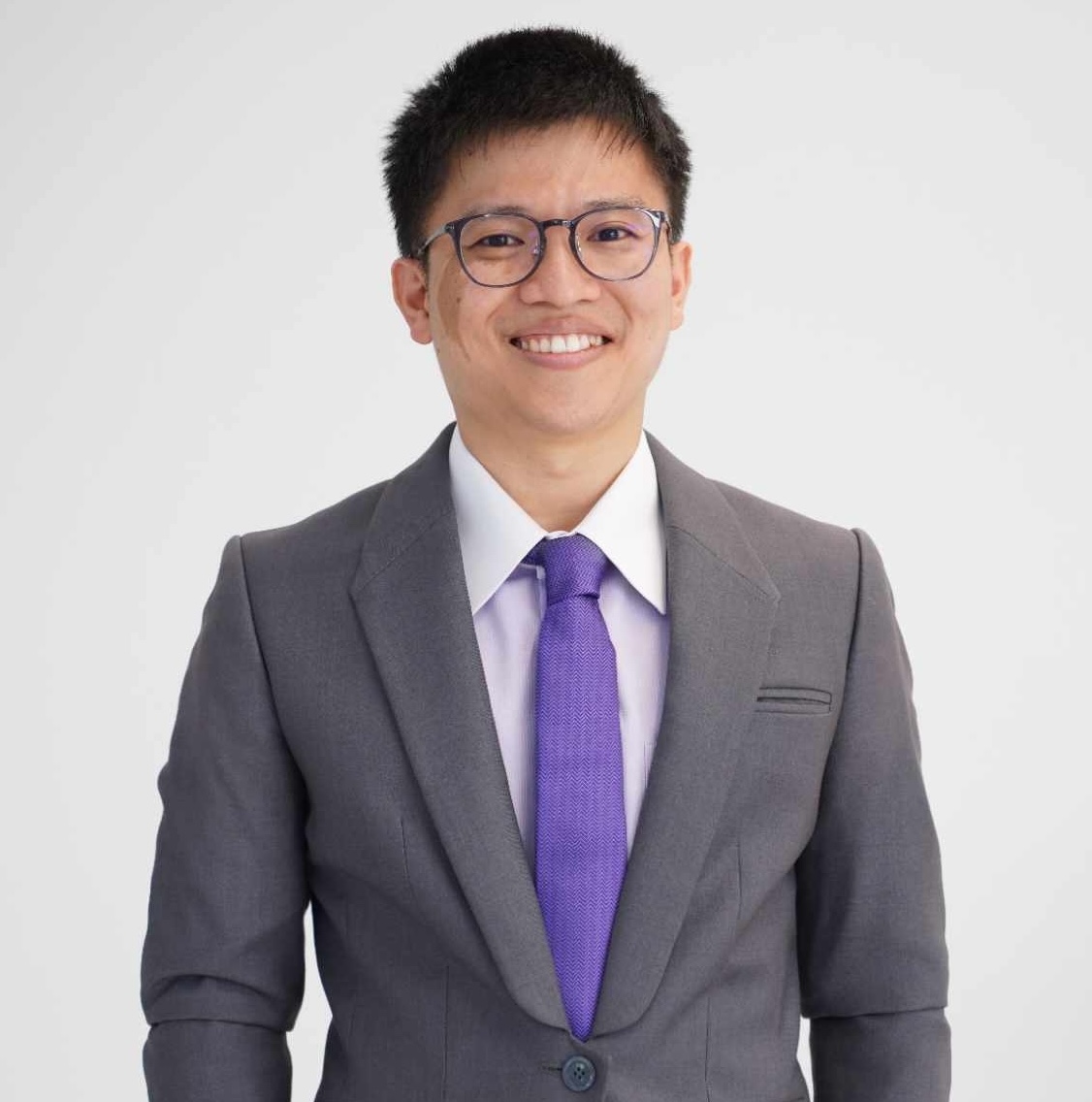}}]{Theerawit Wilaiprasitporn} is a scientist specializing in medical AI and a passionate advocate for deep tech startups; founded Interfaces (AI in Health Research Team at VISTEC) and SensAI (AI-Driven Anomaly Sensing for Better Health). His efforts have played a crucial role in building remote health monitoring systems, benefitting over 30,000 people during the COVID-19 pandemic. This earned him a 2022 IEEE R10 Humanitarian Technology Activities Outstanding Volunteer Award nomination. Dr. Theerawit serves as a supervisor for postgraduate students at VISTEC, guiding research and development efforts. He is actively involved in establishing a university spinoff company that will create awareness of Thailand's potential as a hub for research and development, contributing to the sustainable growth of the Thai economy and industry. Dr Theerawit remains active in the Institute of Electrical and Electronics Engineers (IEEE), further solidifying his impact on advancing technology for the benefit of humanity.
\end{IEEEbiography}
\vspace{-8mm}
\end{document}